\pdfoutput=1
\documentclass[sigconf, nonacm]{acmart}

\usepackage{pvldb}

\renewcommand\vldbdoi{XX.XX/XXX.XX}
\renewcommand\vldbpages{XXX-XXX}
\renewcommand\vldbavailabilityurl{}

\usepackage{amsmath}
\usepackage{tikz}
\usepackage{algorithm}
\usepackage{algpseudocode}
\usepackage{placeins}
\usepackage{stfloats}
\usepackage{colortbl}
\usepackage{multirow}
\usepackage{tabularx}
\usepackage{enumitem}
\newcolumntype{R}{>{\raggedleft\arraybackslash}X}
\definecolor{graywoea}{gray}{0.92}
\definecolor{grayinspire}{gray}{0.81}

\renewcommand{\topfraction}{0.92}
\renewcommand{\dbltopfraction}{0.92}
\renewcommand{\textfraction}{0.06}
\AtBeginDocument{%
    \setlength{\abovedisplayskip}{4.5pt plus 1.5pt minus 1.5pt}%
    \setlength{\belowdisplayskip}{4.5pt plus 1.5pt minus 1.5pt}%
    \setlength{\abovedisplayshortskip}{2.5pt plus 1pt minus 1pt}%
    \setlength{\belowdisplayshortskip}{3.5pt plus 1pt minus 1pt}%
}
\providecommand{\papertitle}{From Shared Patterns to Individual Evidence: Two-Stage Reasoning for Emerging Entities in Temporal Knowledge Graphs}
\providecommand{\method}{\textsc{INSPIRE}}
\providecommand{\component}[1]{\textsc{#1}}
\providecommand{\ablation}[1]{\textnormal{w/o}~\component{#1}}
\providecommand{\predictoronly}{\method{} \ablation{EA}}

\begin{document}
\title{\papertitle}

\author{Jinchuan Zhang$^1$, Lei Gao$^1$, Hongyu Jiang$^1$, Chong Mu$^1$, Chenang Cai$^1$, Ke Yan$^1$$^\dagger$}
\affiliation{%
    \institution{$^1$School of Computer Science and Engineering,\\
        University of Electronic Science and Technology of China, Chengdu, China\\
        Email: \{jc.zhang, muchong, kyan\}@uestc.edu.cn,\\
        \{leigao, hyjiang\}@std.uestc.edu.cn,\\
        \{caichenang\}@hotmail.com\\
        $^\dagger$Corresponding author: Ke Yan.}
    \country{}
}
\renewcommand{\shortauthors}{Jinchuan Zhang et al.}
\renewcommand{\vldbauthors}{Jinchuan Zhang, Lei Gao, Hongyu Jiang, Chong Mu, Chenang Cai, and Ke Yan}

\begin{abstract}
    Temporal knowledge graphs (TKGs) continually introduce emerging entities whose interaction histories evolve from zero-shot debuts into sparse, bursty few-shot trajectories.
However, current paradigms are fragmented: pattern transfer methods lack catalog-wide event memory, whereas online memory models suffer from endpoint role asymmetry that degrades unseen answer ranking.
Furthermore, sparse early interactions exhibit both long-range dispersion and local redundancy under skewed distribution.
To address these challenges, we propose \method{} (\emph{Integrating Shared Patterns and Individual Evidence for Reasoning}), a two-stage forecasting framework over a known catalog.
In Stage I, \method{} learns an availability-aware shared-pattern predictor that fuses event-updated memory across the full candidate catalog while applying coordinated query-side and catalog-side masking.
Subsequently, Stage II freezes this predictor and optimizes a lightweight evidence adapter (\component{EA}) combining textual semantic priors with correlation-weighted long-range event statistics under support-stratified and role-balanced co-answer supervision with prior-odds debiasing and soft residual alignment.
In addition, we decompose evaluation by endpoint roles and track longitudinal emerging-answer cohorts via query micro and answer entity macro metrics.
Across four benchmarks based on the ICEWS and GDELT event datasets, \method{} improves Emerging MRR by $16.73\%$--$88.83\%$ over the strongest baselines.
Relative to paired Stage I predictors, \component{EA} further increases Emerging MRR by $8.28\%$--$32.63\%$, Answer-new MRR by $7.81\%$--$41.24\%$, and cohort macro MRR by $9.64\%$--$39.94\%$, gains attributed to semantic priors and candidate score competition suppressing mismatched established competitors without zero-support prior shortcuts.
The official code is publicly available at \url{https://github.com/KcAcoZhang/INSPIRE}.

\end{abstract}

\maketitle

\vldbtopmatter

\section{Introduction}
\label{sec:introduction}

Real-world relational event streams, ranging from geopolitical repositories such as ICEWS and GDELT to evolving distributed graph systems, are modeled as temporal knowledge graphs (TKGs) of timestamped quadruples $(h,r,o,t)$~\cite{ma2026reldins,song2025ltge,tian2021kgsurvey,tian2026hybrid,zhao2026transfir,zheng2022distributed,zheng2025metacomputing}.
Temporal extrapolation forecasts future links $(h,r,?,t)$ from strictly past facts prior to $t$ by ranking candidate entities across a catalog~\cite{jin2020renet,li2021regcn,zhang2025booster}.
While recent models capture snapshot evolution, multigraph dependencies, logical rules, and histories relevant to the query~\cite{jin2020renet,li2021regcn,liu2022tlogic,sun2025expath,zhang2024multigraph,zhang2025hisres}, continuous streams constantly introduce new actors whose histories are initially absent and accumulate sparsely~\cite{wang2022metatkgr,lee2023ingram,lin2026ontotkge,zhang2025booster}.
Consequently, forecasting in dynamic streams hinges on the \emph{few-shot temporal reasoning} problem: predicting reliably over entities that transition from zero-shot debuts to sparse, unevenly observed trajectories.

Empirical profiling across four standard benchmarks exposes a sharp structural tension, as illustrated in \autoref{fig:emergence-example}(c) (with complete statistics in the supplemental material).
At the identity level, entities first observed during the test interval constitute $13.33\%$--$25.20\%$ of test-active actors; however, at the traffic level, queries touching a zero-support endpoint represent only $0.166\%$--$4.54\%$ of evaluated queries.
Furthermore, emergence extends well beyond an entity's first appearance: on ICEWS05--15, $83.32\%$ of test queries targeting test-emergent answer entities occur after debut under limited positive support, whereas on ICEWS18 $70.35\%$ remain at zero support.
Hence, emerging-entity reasoning spans a continuous spectrum from zero-shot priors to sparse few-shot evidence.

To illustrate this dilemma concretely, \autoref{fig:emergence-example}(a)--(b) tracks the emerging entity \emph{Takata} in ICEWS14, which is preregistered in the candidate catalog with textual metadata prior to its first recorded interaction~\cite{zhao2026transfir}.
At its debut ($t=310$), \emph{Takata} has zero graph history ($n_o=0$).
Whereas querying an unseen head can transfer interaction patterns from semantically related actors, ranking \emph{Takata} as the ground-truth answer to (\emph{Newspaper} (\emph{Japan}), \emph{Make statement}, ?, $310$) requires it to outrank thousands of richly observed catalog competitors.
Subsequently, before the follow-up query (\emph{Japan}, \emph{Make statement}, ?, $332$), \emph{Takata} accumulates only six sparse facts spanning $t=310$--$330$: two early isolated interactions ($t=310,317$), an independent statement from \emph{Jay Rockefeller} ($t=328$), and repeated contacts from the U.S.\ government including two identical \emph{Demand} events ($t=323,330$).
Because short sliding windows discard those earlier sparse facts while independent aggregation overcounts the repeated demands, both heuristics skew \emph{Takata}'s latent state.

\begin{figure}[t]
    \centering
    \includegraphics[width=0.95\columnwidth]{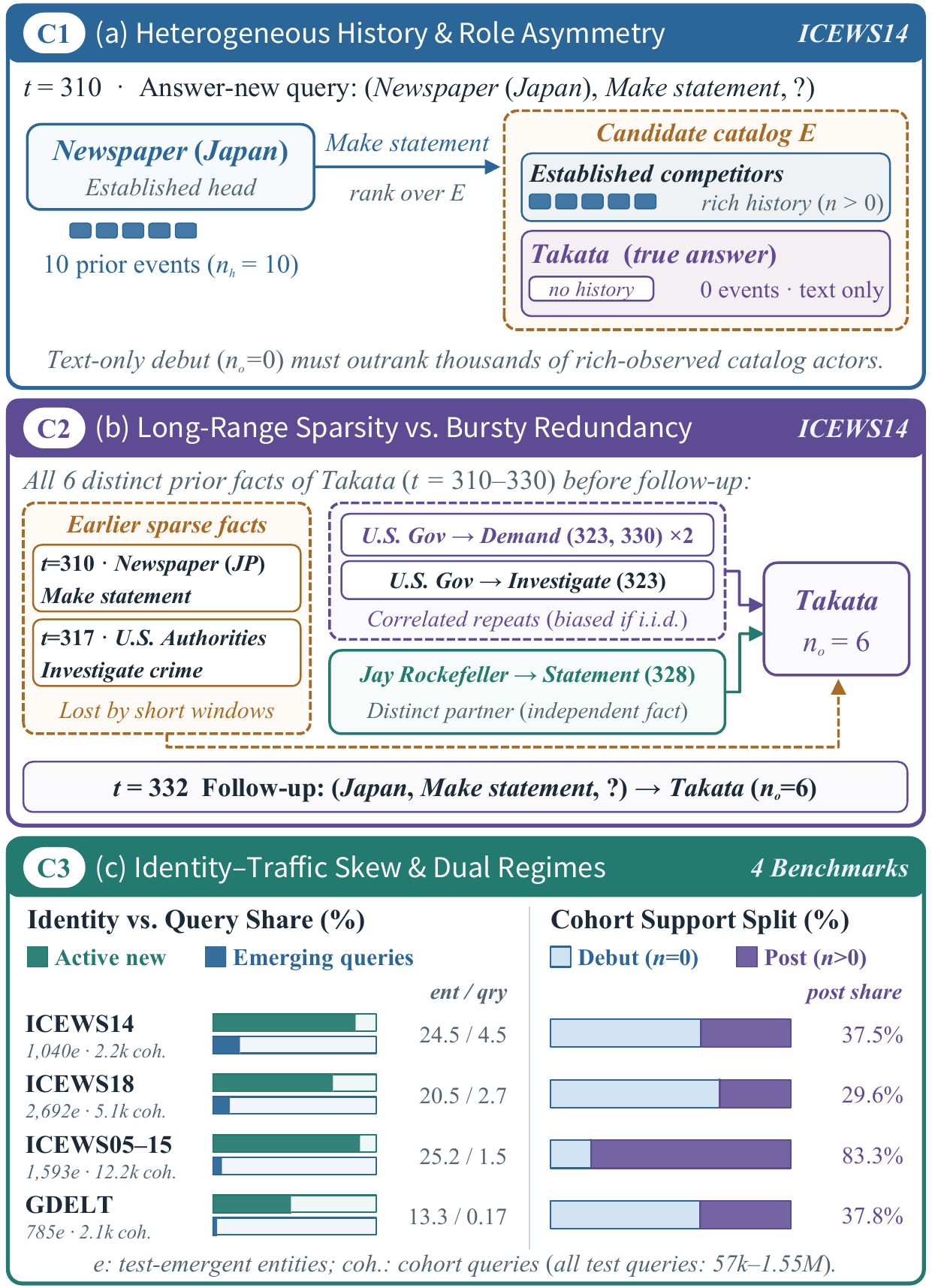}
    \caption{Motivation from emergence events across ICEWS and GDELT benchmarks: (a) \textbf{C1} zero-support debut ($t=310$) competing against richly observed actors; (b) \textbf{C2} sparse early facts ($t=310,317$), correlated U.S.\ Gov repeats ($t=323,330$), and an independent partner ($t=328$); (c) \textbf{C3} identity--traffic skew and debut vs.\ post-debut cohort splits.}
    \Description{Three stacked panels illustrating challenges C1, C2, and C3 using the ICEWS14 Takata entity trajectory and four-benchmark statistics.}
    \label{fig:emergence-example}
\end{figure}

However, existing temporal inductive methods address only isolated ends of this continuum.
Pattern transfer models such as TransFIR~\cite{zhao2026transfir} map unseen entities to semantic interaction clusters at debut, yet omit event-updated candidate memory across the catalog.
In contrast, online memory models such as AdaTKG~\cite{lee2026adatkg} update per-entity exponential moving average (EMA) states, yet exhibit pronounced endpoint role asymmetry: across all four benchmarks (\autoref{tab:role-cohort-results}), AdaTKG improves unseen query prediction (Query-new) over TransFIR while consistently degrading unseen answer ranking (Answer-new), dropping by $57.6\%$ on GDELT ($0.0801 \to 0.0340$).
Therefore, bridging shared patterns with individual evidence requires resolving three coupled challenges:

\noindent\textbf{C1: Heterogeneous history availability and role asymmetry.}
Within any timestamp batch, query heads and catalog candidates span zero-support newcomers and mature actors.
Restricting memory readout to active query heads leaves competing candidates misaligned, whereas naive memory fusion causes excessive reliance on history when observations are absent.

\noindent\textbf{C2: Long-range sparsity versus bursty evidence redundancy.}
Early interactions are simultaneously sparse across long temporal spans and correlated within short bursts.
Updating a high-capacity pattern encoder with sparse gradients risks overfitting; conversely, independent aggregation lets repetitive bursts overwhelm informative facts.

\noindent\textbf{C3: Long-tail supervision dilution and micro-averaged evaluation.}
Since low-support queries occupy a small fraction of natural batches, standard cross-entropy is dominated by high-support entities and neglects asymmetric endpoint roles (Query-new, Answer-new, Both-new).
In addition, conventional micro averaging over queries masks systematic errors on infrequent emerging identities.

To address the aforementioned challenges, we propose \method{} (\emph{Integrating Shared Patterns and Individual Evidence for Reasoning}), a two-stage framework decoupling low-variance shared-pattern learning from lightweight posterior refinement for individual entities.
In Stage I, \method{} learns an availability-aware shared-pattern predictor that fuses event-updated memory across the entire candidate catalog, inclusive of non-query candidates, while applying coupled query-side and catalog-side availability masking during training (\textbf{C1}).
Subsequently, in Stage II, we freeze the Stage I predictor to prevent sparse overfitting and train a lightweight \emph{Evidence Adaptation} (\component{EA}) module.
Specifically, \component{EA} combines a textual semantic prior with bounded long-range evidence over the full revealed prefix, discounts bursty redundancy via a descriptor Gram matrix covariance model, and adds a strictly positively scaled translational Gaussian compatibility residual to full-catalog scores (\textbf{C2}).
Stage II is trained offline under support-stratified and role-balanced co-answer supervision equipped with training-only prior-odds debiasing across support buckets and soft residual alignment to prevent prior shortcuts induced by balanced losses, whereas evaluation drops the training bias, keeps parameters frozen, and updates persistent states strictly after each timestamp is scored (\textbf{C3}).

Across ICEWS14, ICEWS18, ICEWS05--15, and GDELT, \method{} attains Emerging MRRs of $0.2264$, $0.1877$, $0.2587$, and $0.1623$, improving over the strongest baseline by $34.48\%$, $32.15\%$, $16.73\%$, and $88.83\%$.
Relative to paired Stage I predictors (\predictoronly{}), \component{EA} further increases Emerging MRR by $8.28\%$--$32.63\%$, Query-new MRR by $9.29\%$--$21.94\%$, and cohort macro MRR by $9.64\%$--$39.94\%$, while also improving overall test MRR on every dataset.
Notably, \component{EA} achieves substantial calibrated gains ($+7.81\%$--$+41.24\%$) on Answer-new queries where the true answer has zero prior events.
Nevertheless, our functional decomposition resolves this apparent paradox, showing that these gains are attributed jointly to the learned semantic prior and candidate score competition, whereby correlated posterior updates demote mismatched established competitors without inflating a blanket zero-support bias.

Our contributions are threefold:
\begin{enumerate}[leftmargin=*, topsep=2pt, itemsep=1pt, parsep=0pt]
    \item \textbf{Availability-aware shared-pattern learning.} We formulate emerging-entity reasoning across the continuum from zero-shot to few-shot reasoning and introduce catalog-wide gated memory fusion with coordinated query-side and catalog-side masking to resolve role asymmetry.
    \item \textbf{Correlated Bayesian evidence adaptation.} We build a lightweight residual adapter over a frozen predictor combining textual priors, bounded long-range queues, and Gram matrix correlation weighting with balanced supervision.
    \item \textbf{Longitudinal evaluation and mechanism dissection.} We evaluate endpoint roles and emerging-answer cohorts via micro and entity macro metrics on four event benchmarks, isolating how candidate competition drives zero-support gains.
\end{enumerate}

\section{Problem Formulation}
\label{sec:task}

\subsection{Temporal Extrapolation for Emerging Entities}
\label{sec:task-setting}
A TKG is an event stream
$\mathcal{F}\subseteq\mathcal{E}\times\mathcal{R}\times\mathcal{E}\times\mathcal{T}$, where $\mathcal{E}$, $\mathcal{R}$, and $\mathcal{T}$ denote the entity catalog, relation set, and ordered timestamp set.
Each quadruple $(h,r,o,t)\in\mathcal{F}$ records an interaction from head $h$ to tail $o$ under relation $r$ at timestamp $t$.
Let $\mathcal{F}_t$ collect facts at timestamp $t$ and
$\mathcal{H}_{<t}=\bigcup_{\tau<t}\mathcal{F}_\tau$ denote the strictly past revealed prefix.
Given query $q=(h,r,?,t)$ and $\mathcal{H}_{<t}$, temporal extrapolation ranks all catalog candidates $o\in\mathcal{E}$ using past events and fixed textual vectors for entity names $x_e\in\mathbb{R}^{d_x}$.

\noindent\textbf{Consistent problem boundary.}
Following the standard emerging-entity benchmark of TransFIR~\cite{zhao2026transfir} and AdaTKG~\cite{lee2026adatkg}, we retain original timestamps split chronologically into training ($50\%$), validation ($20\%$), and test ($30\%$) intervals, with candidate catalog $\mathcal{E}$ and fixed textual vectors $\{x_e\}_{e\in\mathcal{E}}$ available prior to an entity's first interaction (\emph{emergence}).
At each timestamp $t$, every fact $(h,r,o,t)\in\mathcal{F}_t$ forms a forward tail query $(h,r,?,t)$ (answer $o$) and an inverse head query $(o,r^{-1},?,t)$ (answer $h$).
Let $\mathcal{B}_t^d$ denote the directional query batch for $d\in\{\mathrm{fwd},\mathrm{inv}\}$ and $\mathcal{U}_t^d=\{h:(h,r,t)\in\mathcal{B}_t^d\}$ its active query heads.
Accordingly, both directions are scored over the full catalog $\mathcal{E}$ using only $(h,r,t)\in\mathcal{B}_t^d$, $\mathcal{U}_t^d$, and $\mathcal{H}_{<t}$ strictly before $\mathcal{F}_t$ is revealed; labels are used exclusively for training losses or ranking after scoring.
During evaluation, learned parameters remain frozen while revealed events $\mathcal{F}_t$ subsequently advance persistent states for future timestamps, and raw full-catalog bidirectional ranks are computed under the released history construction pipeline~\cite{zhao2026transfir} across all comparators.

\subsection{Evaluation Protocol}
\label{sec:task-roles}
To trace entities from zero-shot debut through sparse few-shot evolution, we define the observed support of entity $e\in\mathcal{E}$ before timestamp $t$ as
\begin{equation}
    n_e(t)=\sum_{(h,r,o,\tau)\in\mathcal{H}_{<t}}
    \mathbf{1}\!\left[e\in\{h,o\}\right],
    \label{eq:observed-support}
\end{equation}
counting distinct original incident facts strictly before $t$ (self-loops count once, whereas inverse queries add no observations).
Hence, $n_e(t)$ reflects lifetime revealed support independently of finite model history windows.

\noindent\textbf{Asymmetric endpoint roles.}
Queries touching at least one zero-support endpoint decompose into three disjoint roles:
(1) \emph{Query-new} ($n_h(t)=0,\ n_o(t)>0$): an unobserved query head predicts an established answer;
(2) \emph{Answer-new} ($n_h(t)>0,\ n_o(t)=0$): an established head ranks an unobserved answer over $\mathcal{E}$; and
(3) \emph{Both-new} ($n_h(t)=0,\ n_o(t)=0$): both endpoints have zero prior support.
Their union defines the \emph{Emerging} query set.
Although bidirectional evaluation gives Query-new and Answer-new identical query counts, they nevertheless test asymmetric tasks: Query-new initiates prediction from an unseen source, whereas Answer-new requires an unobserved candidate to outrank thousands of established actors.

\noindent\textbf{Longitudinal emerging-answer cohort and metrics.}
Let $\operatorname{first}(e)$ be the earliest timestamp at which $e$ appears in an original fact.
For an evaluation interval $S$ with query set $\mathcal{Q}_S$, the \emph{emerging-answer cohort} $\mathcal{C}_S$ and its longitudinal query set $\mathcal{Q}_{\mathcal{C}_S}$ are
\begin{equation}
    \mathcal{C}_S=\{e\in\mathcal{E}:\operatorname{first}(e)\in S\},\qquad
    \mathcal{Q}_{\mathcal{C}_S}=\{q\in\mathcal{Q}_S:a(q)\in\mathcal{C}_S\},
    \label{eq:answer-cohort}
\end{equation}
where $a(q)$ is the ground-truth answer of $q$.
Consequently, $\mathcal{Q}_{\mathcal{C}_S}$ tracks every answer occurrence of test-emergent identities across debut ($n_o(t)=0$) and post-emergence support strata ($1$, $2$--$5$, $6$--$10$, $>10$).

For any query collection $\mathcal{Q}$, let $\mathcal{E}_{\mathcal{Q}}=\{a(q):q\in\mathcal{Q}\}$ and $\mathcal{Q}_e=\{q\in\mathcal{Q}:a(q)=e\}$.
To prevent frequent actors from dominating aggregate scores, we pair query micro MRR ($\operatorname{MRR}_{\mathrm{micro}}$) with answer entity macro MRR ($\operatorname{MRR}_{\mathrm{macro}}$) over raw full-catalog ranks $\operatorname{rank}(q)$:
\begin{equation}
    \begin{aligned}
        \operatorname{MRR}_{\mathrm{micro}}(\mathcal{Q})
         & =\frac{1}{|\mathcal{Q}|}\sum_{q\in\mathcal{Q}}\frac{1}{\operatorname{rank}(q)}, \\
        \operatorname{MRR}_{\mathrm{macro}}(\mathcal{Q})
         & =\frac{1}{|\mathcal{E}_{\mathcal{Q}}|}
        \sum_{e\in\mathcal{E}_{\mathcal{Q}}}\frac{1}{|\mathcal{Q}_e|}
        \sum_{q\in\mathcal{Q}_e}\frac{1}{\operatorname{rank}(q)}.
    \end{aligned}
    \label{eq:cohort-mrr}
\end{equation}
While $\operatorname{MRR}_{\mathrm{micro}}$ weights queries equally, $\operatorname{MRR}_{\mathrm{macro}}$ weights each distinct answer identity equally over its full trajectory in $\mathcal{Q}$, thereby exposing whether gains extend to infrequent emerging actors.

\subsection{Empirical Diagnosis and Research Questions}
\label{sec:task-motivation}
\autoref{fig:emergence-example}(c) reveals two defining properties of few-shot emergence across the four benchmarks.
First, regarding identity--traffic skew, test-emergent entities account for $13.33\%$--$25.20\%$ of active identities, whereas Emerging queries occupy only $0.166\%$--$4.54\%$ of directed test records; hence, standard training losses severely underweight emerging identities.
Second, regarding dual zero-shot and few-shot regimes, post-emergence workload varies markedly across horizons: on the 11-year ICEWS05--15 stream, $10{,}206$ of $12{,}249$ cohort queries ($83.32\%$) occur after the emerging answer acquires positive support, whereas on ICEWS18 zero-support debuts still comprise $70.35\%$ of cohort queries.

\noindent\textbf{Core hypothesis and research questions.}
Therefore, we hypothesize that decoupling availability-aware shared patterns from correlation-weighted Bayesian adaptation resolves role asymmetry and bursty few-shot sparsity without overfitting high-capacity encoders.
Accordingly, we investigate this hypothesis through six research questions:
\begin{itemize}[leftmargin=*, nosep, topsep=1.5pt]
    \item \textbf{RQ1 (Effectiveness and Role Asymmetry):} Does two-stage decoupling improve Emerging ranking and repair the Answer-new degradation of existing temporal inductive baselines?
    \item \textbf{RQ2 (Availability-Aware Shared Patterns):} How do gated memory fusion, full-catalog readout, and availability masking shape Stage I and support Stage II adaptation?
    \item \textbf{RQ3 (Correlated Evidence and Role Balance):} What distinct contributions come from long-range queues, correlation weighting, and role-balanced supervision?
    \item \textbf{RQ4 (Longitudinal Evolution):} Do gains persist across natural support strata ($0$, $1$, $2$--$5$, $6$--$10$, $>10$) and across distinct identities?
    \item \textbf{RQ5 (Zero-Support Paradox):} Through which inference pathways does individual evidence boost zero-support Answer-new targets?
    \item \textbf{RQ6 (Robustness and Efficiency):} How sensitive is \method{} to masking and evidence hyperparameters, and what overhead does Stage II incur?
\end{itemize}

\section{Methodology}
\label{sec:method}

\subsection{Overview and Two-Stage Decoupling}
\label{sec:method-overview}
\begin{figure*}[t]
    \centering
    \includegraphics[width=0.96\textwidth]{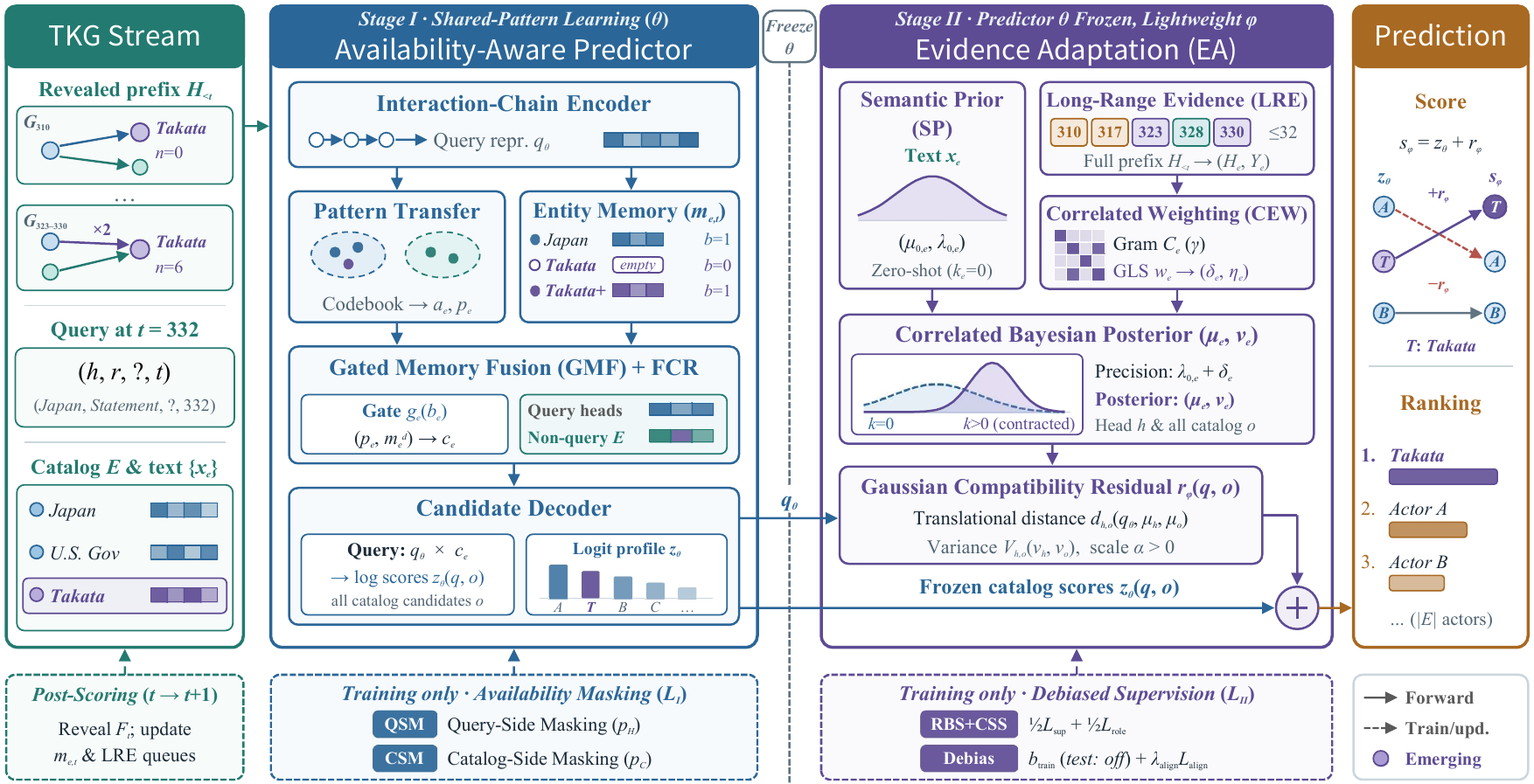}
    \caption{\method{} two-stage framework. Stage I learns an availability-aware predictor ($\theta$) with \component{GMF}, \component{FCR}, and training-only \component{QSM}/\component{CSM}; Stage II freezes $\theta$ and trains lightweight \component{EA} ($\phi$) combining \component{SP}, \component{LRE} ($K=32$), and \component{CEW} into correlated posteriors and a Gaussian residual ($\alpha>0$) under debiased \component{RBS}/\component{CSS} and alignment.}
    \Description{Left-to-right four-zone framework diagram from TKG stream snapshots through Stage I availability-aware predictor across a freeze boundary via two straight parallel arrows into Stage II Evidence Adaptation and full-catalog score competition.}
    \label{fig:architecture}
\end{figure*}

To unify zero-shot pattern transfer with sparse, correlated few-shot evidence, \method{} separates low-variance shared structural priors from posterior refinement for individual entities (\autoref{fig:architecture}).
Given catalog textual vectors $x_e\in\mathbb{R}^{d_x}$ and predictor width $d_\theta$, training proceeds sequentially on the same chronological training split.
In Stage I, \method{} learns an availability-aware shared-pattern predictor ($\theta$) that maps query $q=(h,r,?,t)$ to a vector $\mathbf{q}_\theta\in\mathbb{R}^{d_\theta}$ conditioned on the relation and full-catalog log scores $z_\theta(q,o)\equiv z_\theta(h,r,o,t)$, integrating \emph{Gated Memory Fusion} (\component{GMF}), \emph{Full-Catalog Readout} (\component{FCR}), \emph{Query-Side Masking} (\component{QSM}), and \emph{Catalog-Side Masking} (\component{CSM}).
Subsequently, Stage II freezes the predictor $\theta$ selected on validation, together with all external entity and relation tensors, and optimizes a lightweight Evidence Adaptation (\component{EA}) module ($\phi$) combining a \emph{Semantic Prior} (\component{SP}), \emph{Long-Range Evidence} (\component{LRE}), and \emph{Correlated Evidence Weighting} (\component{CEW}) under support-stratified and \emph{Role-Balanced Supervision} (\component{RBS}) with \emph{Co-Answer Set Supervision} (\component{CSS}).

\noindent\textbf{Rationale for decoupling.}
Since established entities generate over $95\%$ of natural stream traffic, end-to-end joint training invites dense gradients to drown out rare newcomers or causes high-capacity encoders to overfit sparse residuals.
Therefore, freezing $\theta$ anchors a stable global geometry while confining $\phi$ to a regularized residual space.
During evaluation, $(\theta,\phi)$ remain fixed while revealed facts $\mathcal{F}_t$ advance persistent states strictly after timestamp $t$ is scored.

\subsection{Stage I: Availability-Aware Predictor}
\label{sec:method-patterns}
Stage I builds on TransFIR's interaction chain encoder, discrete behavioral codebook, pattern transfer, and candidate decoder~\cite{zhao2026transfir}.
For each query $q_j=(h_j,r_j,?,t)\in\mathcal{B}_t^d$, the encoder maps $h_j$'s recent chain and $r_j$ to $\mathbf{q}_{\theta,j}\in\mathbb{R}^{d_\theta}$, while pattern transfer produces a catalog table $\mathbf{a}_e$ conditioned on the query and an initial pattern vector $\mathbf{p}_e$ for every $e\in\mathcal{E}$.

\noindent\textbf{Event-updated entity memory.}
Following adaptive memory~\cite{lee2026adatkg}, each entity maintains a state $\mathbf{m}_{e,t}\in\mathbb{R}^{d_\theta}$ updated at each timestamp boundary via a learned retention gate $\rho=\sigma(\rho_{\mathrm{raw}})$:
$\mathbf{m}_e\leftarrow\rho\mathbf{m}_e+(1-\rho)\bar{\mathbf{q}}_{e,t}$,
where $\bar{\mathbf{q}}_{e,t}$ averages detached query vectors staged for head $e$ across both directions at $t$ (inactive entities retain $\mathbf{m}_e$).
Committed states become readable only strictly after $t$.
During scoring of $\mathcal{B}_t^d$, active query heads $e\in\mathcal{U}_t^d$ incorporate a causal current batch average $\bar{\mathbf{q}}_e^d$ alongside the past memory view $\bar{\mathbf{m}}_{e,t}$ (equal to $\mathbf{m}_{e,t}$ at evaluation and masked during training):
\begin{equation}
    \widetilde{\mathbf{m}}_e^d=
    \begin{cases}
        \rho\bar{\mathbf{m}}_{e,t}+(1-\rho)\bar{\mathbf{q}}_e^d,
                                & e\in\mathcal{U}_t^d,    \\
        \bar{\mathbf{m}}_{e,t}, & e\notin\mathcal{U}_t^d.
    \end{cases}
    \label{eq:memory-view}
\end{equation}

\noindent\textbf{Gated Memory Fusion (\component{GMF}) and Full-Catalog Readout (\component{FCR}).}
However, restricting memory readout to $\mathcal{U}_t^d$ leaves competing non-query candidates in stale representations and exacerbates Answer-new degradation.
Let $b_e\in\{0,1\}$ indicate whether entity $e$ has unmasked past memory or a current signal in $\mathcal{U}_t^d$.
Accordingly, \component{GMF} applies a coordinate-wise fusion gate across the entire catalog $\mathcal{E}$:
\begin{equation}
    \begin{aligned}
        \mathbf{g}_e & =b_e\,\sigma\!\left(W_g[\mathbf{a}_e;
        \widetilde{\mathbf{m}}_e^d]+\mathbf{b}_g\right),     \\
        \mathbf{c}_e & =\operatorname{norm}_2\!\left(
        (\mathbf{1}_{d_\theta}-\mathbf{g}_e)\odot\mathbf{p}_e+
        \mathbf{g}_e\odot\widetilde{\mathbf{m}}_e^d\right),
    \end{aligned}
    \label{eq:catalog-fusion}
\end{equation}
where $\mathbf{1}_{d_\theta}\in\mathbb{R}^{d_\theta}$ is the vector of ones, $\operatorname{norm}_2$ is $\ell_2$ normalization, and $\odot$ is element-wise product.
Subsequently, the candidate decoder computes full-catalog log-softmax scores $z_\theta(q,o)$ from $\{\mathbf{c}_e:e\in\mathcal{E}\}$.
Through \component{FCR}, every previously observed non-query candidate $e\in\mathcal{E}\setminus\mathcal{U}_t^d$ injects $\bar{\mathbf{m}}_{e,t}$ into $\mathbf{c}_e$, thereby calibrating full-catalog competition.

\noindent\textbf{Coordinated availability masking (\component{QSM} and \component{CSM}).}
Since historical training batches are dominated by observed entities, unmasked training causes $\mathbf{g}_e$ to rely excessively on memory and underperform on zero-support newcomers.
To mitigate this mismatch, we draw independent Bernoulli masks $A_j\sim\operatorname{Bernoulli}(p_H)$ per query row and $B_e\sim\operatorname{Bernoulli}(p_C)$ per catalog entity for each training batch.
Specifically, Query-Side Masking (\component{QSM}) empties row $j$'s interaction chain and masks head $h_j$'s past memory when $A_j=1$ (simulating Query-new), whereas Catalog-Side Masking (\component{CSM}) masks the past memory of any entity with $B_e=1$ (simulating Answer-new competition):
\begin{equation}
    \mathcal{M}_t^d=\{h_j:A_j=1\}\cup\{e:B_e=1\},\qquad
    \bar{\mathbf{m}}_{e,t}=\mathbf{1}[e\notin\mathcal{M}_t^d]\mathbf{m}_{e,t},
    \label{eq:availability-masks}
\end{equation}
clearing the past memory availability bit for $e\in\mathcal{M}_t^d$.
Chain masking acts per row while memory masking applies to $\mathcal{M}_t^d$, after which \autoref{eq:memory-view} is evaluated and staged.
Both masks are disabled at evaluation, with $(p_H,p_C)$ selected on validation (\autoref{sec:experimental-setup}).
Stage I minimizes negative log-likelihood jointly with the codebook loss $\mathcal{L}_{\mathrm{code}}$:
\begin{equation}
    \mathcal{L}_{\mathrm{I}}=
    -\frac{1}{|\mathcal{B}_t^d|}\sum_{j} z_\theta(q_j,o_j)
    +\beta\mathcal{L}_{\mathrm{code}},\qquad \beta=0.1.
    \label{eq:predictor-objective}
\end{equation}

\subsection{Stage II: Evidence Adaptation}
\label{sec:method-evidence}
Although Stage I builds a calibrated shared backbone, EMA memory compresses history into a single vector without separating long-range sparse facts from short-term bursts.
Recalling our aforementioned \emph{Takata} example (\autoref{fig:emergence-example}(b)), retaining early isolated interactions ($t=310,317$) alongside an independent partner ($t=328$) while discounting repeated U.S.\ government demands ($t=323,330$) requires explicit statistics at the event level.
Hence, Stage II (\component{EA}) models each entity's latent state in a compact space $\mathbb{R}^D$ ($D=64$) via correlated Bayesian updating.

\noindent\textbf{Semantic Prior (\component{SP}) and Long-Range Evidence (\component{LRE}).}
To support zero-support entities ($k_e=0$) seamlessly, \component{SP} maps $x_e$ to an isotropic Gaussian prior mean $\boldsymbol{\mu}_{0,e}\in\mathbb{R}^D$ and precision $\lambda_{0,e}>0$:
\begin{equation}
    \boldsymbol{\mu}_{0,e}=W_0x_e,\qquad
    \lambda_{0,e}=\operatorname{softplus}(\mathbf{a}_0^\top x_e+b_0)+\varepsilon,
    \label{eq:semantic-prior}
\end{equation}
with $\varepsilon=10^{-6}$ and parameters $(W_0,\mathbf{a}_0,b_0)$ shared across $\mathcal{E}$.
Furthermore, to avoid losing early sparse events to short temporal windows, \component{LRE} retrieves the latest $k_e\leq K$ ($K=32$) distinct original incident facts of $e$ from the entire revealed prefix $\mathcal{H}_{<t}$ with unit age weights.
For retained event $i$, let $\mathbf{r}_i\in\mathbb{R}^{d_\theta}$ be the frozen directed relation vector, $x_{n_i}\in\mathbb{R}^{d_x}$ the counterpart's text vector, and $\mathbf{d}_i\in\{(1,0),(0,1)\}$ the direction indicator (incoming events use inverse relations with $(0,1)$; self-loops contribute one outgoing record).
We decouple frozen normalized descriptors $\mathbf{u}_i$ from learned surrogate observations $\mathbf{y}_i\in\mathbb{R}^D$:
\begin{equation}
    \mathbf{u}_i=\frac{[\mathbf{r}_i;x_{n_i};\mathbf{d}_i]}{\|[\mathbf{r}_i;x_{n_i};\mathbf{d}_i]\|_2},\qquad
    \mathbf{y}_i=W_R\mathbf{r}_i+W_Nx_{n_i}+W_D\mathbf{d}_i\in\mathbb{R}^{D},
    \label{eq:event-descriptor}
\end{equation}
stacking rows into $H_e\in\mathbb{R}^{k_e\times(d_\theta+d_x+2)}$ and $Y_e\in\mathbb{R}^{k_e\times D}$.

\noindent\textbf{Correlated Evidence Weighting (\component{CEW}).}
Standard Bayesian Context Aggregation~\cite{volpp2021bayesian} assumes independent observations, causing repeated bursts to inflate precision linearly.
On the contrary, \component{CEW} models dependence between events via the descriptor Gram matrix $H_eH_e^\top$ and computes generalized least-squares weights $\mathbf{w}_e$:
\begin{equation}
    \begin{aligned}
        C_e                 & =I_{k_e}+\gamma H_eH_e^\top,        &
        \mathbf{w}_e        & =C_e^{-1}\mathbf{1}_{k_e},            \\
        \delta_e            & =\mathbf{1}_{k_e}^\top\mathbf{w}_e, &
        \boldsymbol{\eta}_e & =Y_e^\top\mathbf{w}_e,
    \end{aligned}
    \label{eq:correlated-statistics}
\end{equation}
where $\mathbf{1}_{k_e}\in\mathbb{R}^{k_e}$ denotes the $k_e$-dimensional vector of ones and $\gamma=0.5$.
Since $C_e\succ 0$, a Cholesky solve computes $\mathbf{w}_e$ stably; $\delta_e>0$ gives the correlation-adjusted effective precision (entries of $\mathbf{w}_e$ may be signed), and $\boldsymbol{\eta}_e$ aggregates weighted evidence.
Under prior $\boldsymbol{\xi}_e\sim\mathcal{N}(\boldsymbol{\mu}_{0,e},\lambda_{0,e}^{-1}I_D)$ and correlated observations $\mathbf{y}_i=\boldsymbol{\xi}_e+\boldsymbol{\epsilon}_i$ with $\operatorname{Cov}(\boldsymbol{\epsilon}_i,\boldsymbol{\epsilon}_j)=(C_e)_{ij}I_D$, the posterior variance and mean are
\begin{equation}
    v_e=(\lambda_{0,e}+\delta_e)^{-1},\qquad
    \boldsymbol{\mu}_e=v_e\left(\lambda_{0,e}\boldsymbol{\mu}_{0,e}
    +\boldsymbol{\eta}_e\right).
    \label{eq:entity-posterior}
\end{equation}
When $k_e=0$, $\delta_e=0$ and $\boldsymbol{\eta}_e=\mathbf{0}$ recover $(\boldsymbol{\mu}_{0,e},\lambda_{0,e}^{-1})$ exactly.
To illustrate this effect, for $k$ identical bursty events ($H_eH_e^\top=\mathbf{1}_k\mathbf{1}_k^\top$), \component{CEW} yields a saturating precision $\delta_e=k/(1+\gamma k)<\gamma^{-1}$, whereas an independent control of matched marginal variance $(1+\gamma)I_k$ (\ablation{CEW}) grows linearly as $k/(1+\gamma)$; in contrast, $k$ orthogonal events across diverse partners accumulate precision nearly linearly.

\noindent\textbf{Translational Gaussian ranking and candidate competition.}
Projecting $\mathbf{q}_\theta$ via $W_Q\in\mathbb{R}^{D\times d_\theta}$, we compute the squared distance
\begin{equation}
    d_{h,o}=\left\|W_Q\mathbf{q}_\theta+
    \boldsymbol{\mu}_h-\boldsymbol{\mu}_o\right\|_2^2.
    \label{eq:adapter-distance}
\end{equation}
Distinct endpoints ($h\neq o$) sum posterior variances $v_h+v_o$, whereas identical endpoints ($h=o$) cancel the shared latent state.
With learned residual scale $\ell=\operatorname{softplus}(\ell_{\mathrm{raw}})+\varepsilon$, we define
\begin{equation}
    V_{h,o}=\begin{cases}
        \ell,         & h=o,     \\
        \ell+v_h+v_o, & h\neq o,
    \end{cases}
    \label{eq:adapter-variance}
\end{equation}
and add the marginal Gaussian compatibility residual $r_\phi(q,o)$ to the frozen predictor:
\begin{equation}
    r_\phi(q,o)=-\frac{\alpha}{2}
    \left(\frac{d_{h,o}}{V_{h,o}}+D\log V_{h,o}\right),\quad
    s_\phi(q,o)=z_\theta(q,o)+r_\phi(q,o),
    \label{eq:final-score}
\end{equation}
where $\alpha=\operatorname{softplus}(\alpha_{\mathrm{raw}};\beta_{\alpha})>0$ ($\beta_{\alpha}=10$, initialized to $10^{-3}$) preserves Gaussian log-likelihood sign semantics, preventing sign inversion from turning $-D\log V_{h,o}$ into an uncalibrated uncertainty bonus for zero-support candidates.
Because rankings depend on margins $s_\phi(q,o)-s_\phi(q,o')$, a zero-support true answer $o$ ($k_o=0$) still gains rank through its semantic prior and when observed competitors $o'$ ($k_{o'}>0$) update $\boldsymbol{\mu}_{o'}$ and contract $v_{o'}$, demoting mismatched established distractors via candidate score competition (\autoref{sec:evidence-mechanisms}).

\subsection{Debiased Balanced Supervision}
\label{sec:method-supervision}
Since natural batches strongly favor high-support entities (\autoref{fig:emergence-example}(c)), Stage II balances rare strata and roles via two equally weighted branches.
However, naive balanced cross-entropy introduces an implicit Bayes prior-odds shift $\log(\pi_{\mathrm{bal}}(g)/\pi_{\mathrm{nat}}(g))$ across answer support strata $g\in\{0,\dots,4\}$ ($n_o(t)\in\{0,1,2\text{--}5,6\text{--}10,>10\}$), which an unconstrained adapter can absorb as a blanket score bonus for zero-support candidates.

\noindent\textbf{Training-only prior-odds debiasing.}
Let $g(e,t)\in\{0,\dots,4\}$ denote candidate $e$'s lifetime support stratum before $t$.
To counter the aforementioned prior-odds shift during training only, we add a learnable support bucket bias $\mathbf{b}_{\mathrm{train}}\in\mathbb{R}^5$ anchored at mature entities ($\tilde{b}_g=b_{\mathrm{train},g}-b_{\mathrm{train},4}$, $\tilde{b}_4=0$), initialized to the closed-form training log-odds $\log(\pi_{\mathrm{bal}}(g)/\pi_{\mathrm{nat}}(g))-\log(\pi_{\mathrm{bal}}(4)/\pi_{\mathrm{nat}}(4))$ without weight decay:
\begin{equation}
    s_\phi^{\mathrm{tr}}(q,o)=s_\phi(q,o)+\tilde{b}_{g(o,t)},\qquad
    p_\phi^{\mathrm{tr}}(o\mid q)=\operatorname{softmax}_o\big(s_\phi^{\mathrm{tr}}(q,o)\big).
    \label{eq:train-bias}
\end{equation}
Subsequently, at validation and test time, $\mathbf{b}_{\mathrm{train}}$ is discarded so that ranking relies exclusively on the unbiased scores $s_\phi(q,o)$ (\autoref{eq:final-score}).

\noindent\textbf{Support-stratified and role-balanced branches.}
For any partition $\mathcal{G}$ with nonempty groups $\mathcal{G}^{+}=\{G\in\mathcal{G}:|G|>0\}$, define
\begin{equation}
    \operatorname{Bal}(\ell,\mathcal{G})=
    \frac{1}{|\mathcal{G}^{+}|}\sum_{G\in\mathcal{G}^{+}}
    \frac{1}{|G|}\sum_{j\in G}\ell_j
    \label{eq:balanced-reduction}
\end{equation}
(zero if $\mathcal{G}^{+}=\emptyset$).
Rows in $\mathcal{B}_t^d$ are grouped by lifetime answer support $n_{o_j}(t)$ (\autoref{eq:observed-support}) into five strata $\mathcal{G}_{\mathrm{sup}}$ ($0,1,2\text{--}5,6\text{--}10,>10$) using cross-entropy $\ell_{\mathrm{CE},j}=-\log p_\phi^{\mathrm{tr}}(o_j\mid q_j)$, yielding $\mathcal{L}_{\mathrm{sup}}=\operatorname{Bal}(\ell_{\mathrm{CE}},\mathcal{G}_{\mathrm{sup}})$.
To balance both endpoints, Role-Balanced Supervision (\component{RBS}) partitions eligible low-support rows into four role groups $\mathcal{G}_{\mathrm{role}}$: Query-new, Answer-new, Both-new, and sparse rows observed at both endpoints having $n_{h_j}(t),n_{o_j}(t)>0$ and minimum support $\min(n_{h_j}(t),n_{o_j}(t))\leq 5$.
Let $\mathcal{A}_t^d(h,r)$ collect all distinct co-answers sharing $(h,r)$ in $\mathcal{B}_t^d$ (formed prior to grouping, keeping repeated rows as separate averaging units).
Accordingly, Co-Answer Set Supervision (\component{CSS}) maximizes total mass on $\mathcal{A}_t^d(h_j,r_j)$:
\begin{equation}
    \ell_{\mathrm{CSS},j}=-\log\!\!\sum_{o\in\mathcal{A}_t^d(h_j,r_j)}\!\!
    p_\phi^{\mathrm{tr}}(o\mid q_j),\quad
    \mathcal{L}_{\mathrm{role}}=\operatorname{Bal}(\ell_{\mathrm{CSS}},\mathcal{G}_{\mathrm{role}}).
    \label{eq:role-set-loss}
\end{equation}

\noindent\textbf{Soft residual alignment.}
Since \autoref{eq:train-bias} is invariant to shifting $r_\phi(q,o)$ up by $c$ on zero-support entities $\mathcal{E}_0(t)=\{e:n_e(t)=0\}$ and shifting $\tilde{b}_0$ down by $c$, $\mathbf{b}_{\mathrm{train}}$ alone retains an additive gauge ambiguity.
Writing $\bar{r}_S(q)=\frac{1}{|S|}\sum_{o\in S} r_\phi(q,o)$ over $\mathcal{E}_0(t)$ and $\mathcal{E}_{>0}(t)=\mathcal{E}\setminus\mathcal{E}_0(t)$, we therefore anchor the residual gauge without altering ordering within each stratum:
\begin{equation}
    \mathcal{L}_{\mathrm{align}}=
    \frac{1}{|\mathcal{B}_t^d|}\sum_{q\in\mathcal{B}_t^d}
    \big(\bar{r}_{\mathcal{E}_0(t)}(q)-\bar{r}_{\mathcal{E}_{>0}(t)}(q)\big)^2,
    \label{eq:align-loss}
\end{equation}
whenever both sets are nonempty, consequently yielding the Stage II objective
\begin{equation}
    \mathcal{L}_{\mathrm{II}}=
    \tfrac12\mathcal{L}_{\mathrm{sup}}+
    \tfrac12\mathcal{L}_{\mathrm{role}}+
    \lambda_{\mathrm{align}}\mathcal{L}_{\mathrm{align}}.
    \label{eq:adapter-objective}
\end{equation}
We average $\mathcal{L}_{\mathrm{sup}}$ and $\mathcal{L}_{\mathrm{role}}$ equally ($\tfrac{1}{2}$ each) rather than introducing an extra trade-off hyperparameter specific to each dataset, thereby preserving the cross-entropy loss scale of a single supervision branch while weighting support-stratified and role-balanced gradients equally (removing \component{RBS} drops the second branch).
In addition, the single global alignment weight $\lambda_{\mathrm{align}}=0.2$ is selected strictly on validation splits (\autoref{sec:experimental-setup}).

\begin{algorithm}[t]
    \caption{Two-stage learning and timestamp evaluation}
    \label{alg:inspire}
    \small
    \begin{algorithmic}[1]
        \Require Chronological splits; entity catalog $\mathcal{E}$ with text $\{x_e\}$
        \State \textbf{Stage I:} Optimize $\theta$ under $\mathcal{L}_{\mathrm{I}}$ with \component{GMF}, \component{FCR}, \component{QSM}, \component{CSM}
        \State Select $\theta$ by validation Emerging MRR; freeze predictor
        \For{each Stage II training epoch}
        \State Reset training memory, evidence queues, and support counts
        \For{training timestamps $t$ in chronological order}
        \For{each directional batch $\mathcal{B}_t^d$}
        \State Obtain frozen $z_\theta,\mathbf{q}_\theta$ and stage query signals
        \State Compute $s_\phi(q,o)$ (Eq.~\ref{eq:final-score}) and $s_\phi^{\mathrm{tr}}$ (Eq.~\ref{eq:train-bias}) from $\mathcal{H}_{<t}$
        \State Update $(\phi,\mathbf{b}_{\mathrm{train}})$ via $\mathcal{L}_{\mathrm{II}}$ (Eq.~\ref{eq:adapter-objective})
        \EndFor
        \State Reveal $\mathcal{F}_t$; advance queues, support counts, and memory
        \EndFor
        \State Validate unbiased $s_\phi$ (dropping $\mathbf{b}_{\mathrm{train}}$) and retain best $\phi$
        \EndFor
        \State \textbf{Evaluation:} Fix $(\theta,\phi)$, drop $\mathbf{b}_{\mathrm{train}}$, and initialize prefix states
        \For{evaluation timestamps $t$ in chronological order}
        \State Score both directions from permitted states and $\mathcal{U}_t^d$
        \State Record full-catalog ranks; reveal $\mathcal{F}_t$ and advance states
        \EndFor
    \end{algorithmic}
\end{algorithm}

\subsection{Complexity Analysis}
\label{sec:complexity}
\autoref{alg:inspire} decouples offline training from state updates at timestamp boundaries.
Since descriptors $\mathbf{u}_i$ use frozen embeddings and fixed text, each entity's Cholesky weights $\mathbf{w}_e$, precision $\delta_e$, and weighted raw relation, neighbor, and direction sums are cached and refreshed only when $\mathcal{F}_t$ modifies its queue; hence, by linearity, projecting cached sums via $(W_R,W_N,W_D)$ recovers $\boldsymbol{\eta}_e$ in \autoref{eq:correlated-statistics} without repeated Cholesky solves.
For $a_t$ updated entities and descriptor width $p=d_\theta+d_x+2$, incremental refreshes cost $O(a_t(K^2p+K^3))$, catalog projections cost $O(|\mathcal{E}|pD)$, residual scoring costs $O(|\mathcal{B}_t^d||\mathcal{E}|D)$, and queue storage is $O(|\mathcal{E}|K)$.
In parameter count, Stage II adds only $2D(d_x+d_\theta)+d_x+2D+3$ inference parameters ($197{,}507$ at $d_x=d_\theta=768$; $230{,}275$ at $d_\theta=1024$), plus $5$ training-only support bias scalars discarded before evaluation.
Validation initializes with empty predictor memory and training prefix evidence queues, and testing carries validation states forward continuously.

\section{Experiments}
\label{sec:experiments}

\subsection{Experimental Setup}
\label{sec:experimental-setup}
\begin{table}[!b]
    \centering
    \small
    \caption{Stage I hyperparameters ($8$ heads, $14$-snapshot window, dropout $0.1$). Shared Stage II \component{EA} fixes $D=64$, $K=32$, $\gamma=0.5$, $\alpha>0$, $\lambda_{\mathrm{align}}=0.2$.}
    \label{tab:experimental-settings}
    \renewcommand{\arraystretch}{0.94}
    \begin{tabular*}{\linewidth}{@{\extracolsep{\fill}}lrrrrrr@{}}
        \toprule
        \multicolumn{1}{c}{Dataset} & $d_\theta$ & Layers & Codes & $L_{\max}$ & $p_H$ & $p_C$\\
        \midrule
        ICEWS14 & 768 & 3 & 50 & 30 & .2 & .5\\
        ICEWS18 & 768 & 2 & 50 & 15 & .2 & .5\\
        ICEWS05--15 & 1024 & 3 & 30 & 10 & .4 & .25\\
        GDELT & 768 & 3 & 100 & 30 & .2 & 0\\
        \bottomrule
    \end{tabular*}
\end{table}

\begin{table*}[!t]
    \centering
    \small
    \caption{Emerging-entity link prediction across four datasets (MRR, H@3, H@10). Results for the first three baseline families follow~\cite{zhao2026transfir,lee2026adatkg} (---: unreported MRR); $^\ast$ denotes methods evaluated under our unified raw, full-catalog bidirectional protocol. Bold/underlined mark best/second-best per column; $\Delta$ is \method{}'s relative gain over the strongest baseline.}
    \label{tab:main-results}
    \setlength{\tabcolsep}{2.5pt}
    \renewcommand{\arraystretch}{0.90}
    \begin{tabularx}{\textwidth}{clRRRRRRRRRRRR}
        \toprule
        \multicolumn{2}{c}{\multirow{2}{*}{Method}} & \multicolumn{3}{c}{ICEWS14} & \multicolumn{3}{c}{ICEWS18} & \multicolumn{3}{c}{ICEWS05--15} & \multicolumn{3}{c}{GDELT}                                                                                                                                                                                              \\
        \cmidrule(lr){3-5}\cmidrule(lr){6-8}\cmidrule(lr){9-11}\cmidrule(l){12-14}
                                                    &                             & MRR                         & H@3                             & H@10                      & MRR                & H@3                & H@10               & MRR                & H@3                & H@10               & MRR                & H@3                & H@10               \\
        \midrule
        \multirow{7}{*}{\rotatebox[origin=c]{90}{\footnotesize Graph-based}}
                                                    & CyGNet                      & 0.0111                      & 0.0098                          & 0.0202                    & 0.0031             & 0.0020             & 0.0047             & 0.0031             & 0.0020             & 0.0048             & 0.0067             & 0.0031             & 0.0147             \\
                                                    & RE-GCN                      & 0.1175                      & 0.1263                          & 0.2232                    & 0.0947             & 0.1004             & 0.1797             & 0.0887             & 0.0961             & 0.1803             & 0.0222             & 0.0209             & 0.0393             \\
                                                    & HiSMatch                    & 0.0284                      & 0.0285                          & 0.0418                    & 0.0055             & 0.0058             & 0.0076             & 0.0242             & 0.0238             & 0.0443             & 0.0159             & 0.0141             & 0.0270             \\
                                                    & MGESL                       & 0.0309                      & 0.0361                          & 0.0603                    & 0.0747             & 0.0809             & 0.1031             & 0.1069             & 0.1166             & 0.1563             & 0.0516             & 0.0471             & 0.0815             \\
                                                    & LogCL                       & 0.1354                      & 0.1770                          & 0.2273                    & 0.0903             & 0.1064             & 0.1548             & 0.1917             & 0.2452             & 0.2855             & 0.0473             & 0.0479             & 0.0973             \\
                                                    & HisRES                      & 0.1169                      & 0.1107                          & 0.2132                    & 0.0445             & 0.0434             & 0.0735             & 0.1325             & 0.1332             & 0.1407             & 0.0416             & 0.0737             & 0.0932             \\
                                                    & MLEMKD                      & 0.0685                      & 0.0728                          & 0.1303                    & 0.0402             & 0.0382             & 0.0831             & 0.0833             & 0.0848             & 0.1717             & 0.0229             & 0.0215             & 0.0436             \\
        \midrule
        \multirow{4}{*}{\rotatebox[origin=c]{90}{\scriptsize Path-based}}
                                                    & TLogic                      & 0.0122                      & 0.0107                          & 0.0257                    & 0.0141             & 0.0131             & 0.0262             & 0.0121             & 0.0108             & 0.0285             & 0.0733             & 0.0749             & 0.1131             \\
                                                    & TILP                        & 0.0397                      & 0.0471                          & 0.1114                    & 0.0498             & 0.0669             & 0.1659             & 0.0358             & 0.0374             & 0.1030             & 0.0053             & 0.0025             & 0.0084             \\
                                                    & ECEformer                   & 0.0461                      & 0.0496                          & 0.0915                    & 0.0323             & 0.0680             & 0.0454             & 0.0587             & 0.0642             & 0.1141             & 0.0426             & 0.0410             & 0.0872             \\
                                                    & GenTKG                      & ---                         & 0.0983                          & 0.1919                    & ---                & 0.0540             & 0.1512             & ---                & 0.1105             & 0.1873             & ---                & 0.0734             & 0.1013             \\
        \midrule
        \multirow{5}{*}{\rotatebox[origin=c]{90}{\footnotesize\shortstack[c]{Static                                                                                                                                                                                                                                                                                        \\Inductive}}}
                                                    & CompGCN                     & 0.0682                      & 0.0906                          & 0.1213                    & 0.0638             & 0.0745             & 0.1049             & 0.1885             & 0.2103             & 0.2479             & 0.0472             & 0.0791             & 0.0934             \\
                                                    & ICL                         & 0.0252                      & 0.0261                          & 0.0388                    & 0.0639             & 0.0727             & 0.0938             & 0.0254             & 0.0302             & 0.0373             & 0.0277             & 0.0326             & 0.0362             \\
                                                    & PPT                         & 0.0093                      & 0.1062                          & 0.1716                    & 0.0368             & 0.0386             & 0.0650             & 0.0015             & 0.0005             & 0.0022             & 0.0406             & 0.0425             & 0.0764             \\
                                                    & MorsE                       & 0.0136                      & 0.0074                          & 0.0185                    & 0.0078             & 0.0075             & 0.0126             & 0.0381             & 0.0167             & 0.0439             & 0.0039             & 0.0040             & 0.0152             \\
                                                    & InGram                      & 0.0563                      & 0.0596                          & 0.1138                    & 0.0254             & 0.0265             & 0.0518             & 0.0771             & 0.0793             & 0.1454             & 0.0471             & 0.0430             & 0.0847             \\
        \midrule
                                                    & TransFIR$^\ast$             & 0.1618                      & 0.1895                          & 0.3174                    & 0.1152             & 0.1346             & 0.2324             & 0.2123             & 0.2547             & 0.3780             & 0.0860             & 0.0847             & 0.1914             \\
                                                    & AdaTKG$^\ast$               & 0.1684                      & 0.1941                          & 0.3332                    & 0.1420             & 0.1619             & 0.2676             & 0.2216             & 0.2547             & 0.3822             & 0.0801             & 0.0797             & 0.1667             \\
        \rowcolor{graywoea}\cellcolor{white}        & \method{} w/o EA            & \underline{0.1918}          & \underline{0.2342}              & \underline{0.3552}        & \underline{0.1510} & \underline{0.1789} & \underline{0.2858} & \underline{0.2389} & \underline{0.2810} & \underline{0.3895} & \underline{0.1224} & \underline{0.1265} & \underline{0.2537} \\
        \rowcolor{grayinspire}\cellcolor{white}     & \method{}                   & \textbf{0.2264}             & \textbf{0.2519}                 & \textbf{0.3702}           & \textbf{0.1877}    & \textbf{0.2076}    & \textbf{0.3097}    & \textbf{0.2587}    & \textbf{0.2928}    & \textbf{0.3980}    & \textbf{0.1623}    & \textbf{0.1725}    & \textbf{0.2738}    \\
        \multirow{-5}{*}{\rotatebox[origin=c]{90}{\footnotesize\shortstack[c]{Temporal                                                                                                                                                                                                                                                                                     \\Inductive}}}
                                                    & $\Delta$                    & +34.48\%                    & +29.76\%                        & +11.10\%                  & +32.15\%           & +28.24\%           & +15.70\%           & +16.73\%           & +14.92\%           & +4.12\%            & +88.83\%           & +103.65\%          & +43.03\%           \\
        \bottomrule
    \end{tabularx}
\end{table*}

\noindent\textbf{Datasets.}
We evaluate on four standard TKG benchmarks, namely ICEWS14, ICEWS18, ICEWS05--15, and GDELT, split chronologically into $50\%/20\%/30\%$ train/validation/test intervals following TransFIR~\cite{zhao2026transfir} and AdaTKG~\cite{lee2026adatkg}, which span distinct emergence regimes (\autoref{fig:emergence-example}(c)).
Specifically, ICEWS14 is short and moderately sparse ($7{,}128$ entities, $365$ snapshots, $249$ facts/snapshot), whereas ICEWS18 has the widest catalog ($23{,}033$ entities, $304$ snapshots) and highest debut concentration ($70.35\%$ of cohort queries at support $0$).
In contrast, ICEWS05--15 spans the longest horizon ($10{,}488$ entities, $4{,}017$ snapshots, $115$ facts/snapshot), where $83.32\%$ of cohort queries occur post-debut; finally, GDELT is highly dense ($7{,}691$ entities, $2{,}751$ snapshots, $828$ facts/snapshot) with severe skew ($0.166\%$ emerging-query share).

\noindent\textbf{Baselines.}
We compare against 18 baselines across four families: Graph-based (CyGNet \cite{zhu2021cygnet}, RE-GCN \cite{li2021regcn}, HiSMatch \cite{li2022hismatch}, MGESL \cite{shi2024mgesl}, LogCL \cite{chen2024logcl}, HisRES \cite{zhang2025hisres}, MLEMKD \cite{qian2025mlemkd}), Path-based (TLogic \cite{liu2022tlogic}, TILP \cite{xiong2023tilp}, ECEformer \cite{fang2024eceformer}, GenTKG \cite{liao2024gentkg}), Static Inductive (CompGCN \cite{vashishth2020compgcn}, ICL \cite{lee2023icl}, PPT \cite{xu2023ppt}, MorsE \cite{chen2022morse}, InGram \cite{lee2023ingram}), and Temporal Inductive (TransFIR \cite{zhao2026transfir}, AdaTKG \cite{lee2026adatkg}).

\noindent\textbf{Evaluation protocol.}
Following \autoref{sec:task-setting}, all models perform raw, full-catalog bidirectional ranking over complete directional batches $\mathcal{B}_t^d$ strictly before $\mathcal{F}_t$ is revealed.
Accordingly, consistent with \autoref{sec:task-roles}, we report Emerging MRR/H@3/H@10, endpoint role MRRs (Answer-new, Query-new, Both-new), and emerging-answer cohort $\operatorname{MRR}_{\mathrm{micro}}$ and $\operatorname{MRR}_{\mathrm{macro}}$.

Results for the first three families follow~\cite{zhao2026transfir,lee2026adatkg}. TransFIR$^\ast$ and AdaTKG$^\ast$ are reproduced from their official implementations alongside \predictoronly{} and \method{} under a fixed random seed.
Checkpoints are selected by validation Emerging MRR (Stage II breaks ties by Answer-new, Both-new, and earlier epoch).
Furthermore, to isolate Stage II cleanly, each dataset's frozen Stage I predictor keeps the $(p_H,p_C)$ selected on validation across Stage II ablations, while the single global alignment weight $\lambda_{\mathrm{align}}=0.2$ is chosen strictly on the ICEWS14 and GDELT validation splits over $\{0,0.05,0.20,1.00\}$ prior to evaluation and fixed across all datasets.

\noindent\textbf{Implementation settings.}
\autoref{tab:experimental-settings} lists the predictor widths and masking rates for each dataset alongside shared Stage II hyperparameters ($D=64$, $K=32$, $\gamma=0.5$, $\alpha>0$, $\lambda_{\mathrm{align}}=0.2$, unit age weights).
Stage I optimizes with Adam (lr $10^{-4}$, weight decay $10^{-5}$, $\leq 1{,}000$ epochs), whereas Stage II uses AdamW (lr $3\times10^{-4}$, weight decay $10^{-4}$, $\leq 50$ epochs; training-only support bias $\mathbf{b}_{\mathrm{train}}$ uses lr $3\times10^{-3}$ with zero weight decay and is discarded at evaluation).
Both stages clip gradient norms at $1.0$ with early stopping after $11$ epochs without improvement under fixed random seed $42$ on NVIDIA RTX 4090 GPUs.
The official implementation is publicly available at \url{https://github.com/KcAcoZhang/INSPIRE}.

\begin{table*}[!t]
    \centering
    \small
    \caption{Endpoint role and emerging-answer cohort MRR for Temporal Inductive methods ($^\ast$: reproduced baselines; bold: best per dataset and column). GDELT Both-new has two directed queries (shown to six decimals).}
    \label{tab:role-cohort-results}
    \renewcommand{\arraystretch}{0.90}
    \begin{tabularx}{\textwidth}{clRRRRR}
        \toprule
        \multicolumn{1}{c}{Dataset}                                          & \multicolumn{1}{c}{Method} & Answer-new      & Query-new       & Both-new          & Cohort micro    & Cohort macro    \\
        \midrule
                                                                             & TransFIR$^\ast$            & 0.1119          & 0.2115          & 0.1631            & 0.1256          & 0.0768          \\
                                                                             & AdaTKG$^\ast$              & 0.0956          & 0.2430          & 0.1539            & 0.1167          & 0.0730          \\
        \rowcolor{graywoea}\cellcolor{white}                                 & \method{} w/o EA           & 0.1425          & 0.2413          & 0.1909            & 0.1646          & 0.1027          \\
        \rowcolor{grayinspire}\cellcolor{white}\multirow{-4}{*}{ICEWS14}     & \method{}                  & \textbf{0.1679} & \textbf{0.2797} & \textbf{0.2677}   & \textbf{0.1913} & \textbf{0.1228} \\
        \midrule
                                                                             & TransFIR$^\ast$            & 0.0685          & 0.1624          & 0.1056            & 0.0802          & 0.0486          \\
                                                                             & AdaTKG$^\ast$              & 0.0644          & 0.2224          & 0.0954            & 0.0810          & 0.0514          \\
        \rowcolor{graywoea}\cellcolor{white}                                 & \method{} w/o EA           & 0.1025          & 0.2009          & 0.1263            & 0.1185          & 0.0705          \\
        \rowcolor{grayinspire}\cellcolor{white}\multirow{-4}{*}{ICEWS18}     & \method{}                  & \textbf{0.1291} & \textbf{0.2449} & \textbf{0.2122}   & \textbf{0.1501} & \textbf{0.0963} \\
        \midrule
                                                                             & TransFIR$^\ast$            & 0.1768          & 0.2499          & \textbf{0.1647}   & 0.2508          & 0.1352          \\
                                                                             & AdaTKG$^\ast$              & 0.1667          & 0.2808          & 0.1236            & 0.2284          & 0.1272          \\
        \rowcolor{graywoea}\cellcolor{white}                                 & \method{} w/o EA           & 0.1961          & 0.2859          & 0.1415            & 0.2718          & 0.1442          \\
        \rowcolor{grayinspire}\cellcolor{white}\multirow{-4}{*}{ICEWS05--15} & \method{}                  & \textbf{0.2114} & \textbf{0.3125} & 0.1094            & \textbf{0.2878} & \textbf{0.1581} \\
        \midrule
                                                                             & TransFIR$^\ast$            & 0.0801          & 0.0919          & \textbf{0.000149} & 0.0669          & 0.0508          \\
                                                                             & AdaTKG$^\ast$              & 0.0340          & \textbf{0.1263} & 0.000139          & 0.0363          & 0.0245          \\
        \rowcolor{graywoea}\cellcolor{white}                                 & \method{} w/o EA           & 0.1541          & 0.0909          & 0.000136          & 0.1149          & 0.0909          \\
        \rowcolor{grayinspire}\cellcolor{white}\multirow{-4}{*}{GDELT}       & \method{}                  & \textbf{0.2176} & 0.1072          & 0.000134          & \textbf{0.1639} & \textbf{0.1272} \\
        \bottomrule
    \end{tabularx}
\end{table*}

\subsection{Overall Performance (RQ1)}
\label{sec:main-results}
To answer \textbf{RQ1}, \autoref{tab:main-results} compares \method{} against all 18 baselines on Emerging link prediction, and \autoref{tab:role-cohort-results} further decomposes the four Temporal Inductive models by endpoint role and longitudinal cohort metrics.

\noindent\textbf{Role asymmetry in baselines.}
Transductive graph, path, and static inductive baselines perform poorly on emerging entities since lookup embeddings and static structural transfer cannot track newly active actors as histories arrive.
Furthermore, \autoref{tab:role-cohort-results} exposes a severe endpoint role trade-off between the two strongest temporal inductive baselines: relative to TransFIR$^\ast$, AdaTKG$^\ast$ raises Query-new MRR across all four datasets by $12.3\%$--$37.4\%$, yet consistently degrades Answer-new MRR by $5.7\%$--$14.6\%$ on the three ICEWS datasets and by $57.6\%$ on dense GDELT ($0.0801\to 0.0340$).
This degradation is attributed to AdaTKG updating EMA memory primarily along query heads without calibrating catalog-wide candidate competition, allowing established candidates to drift ahead of unseen answer targets.
Consequently, AdaTKG$^\ast$ also trails TransFIR$^\ast$ in cohort $\operatorname{MRR}_{\mathrm{macro}}$ on three of the four datasets (dropping $51.8\%$ on GDELT).

\noindent\textbf{Two-stage gains across roles and identities.}
On the contrary, \method{} resolves the aforementioned asymmetry progressively across its two stages.
In Stage I, catalog-wide gated fusion (\component{GMF}, \component{FCR}) and availability masking allow \predictoronly{} to surpass every baseline across all metrics in \autoref{tab:main-results} and reverse AdaTKG's Answer-new collapse on all four datasets, improving Answer-new MRR over AdaTKG$^\ast$ by $17.6\%$--$59.3\%$ on ICEWS and by $353.1\%$ on GDELT.
Subsequently, Stage II \component{EA} builds directly on this calibrated backbone, adding a further $8.28\%$--$32.63\%$ relative Emerging MRR gain over \predictoronly{} ($+18.05\%$ on ICEWS14, $+24.33\%$ on ICEWS18, $+8.28\%$ on ICEWS05--15, and $+32.63\%$ on GDELT) to outperform the strongest baseline by $16.73\%$--$88.83\%$ in MRR and up to $103.65\%$ in H@3.
Moreover, these emerging gains do not sacrifice mature entities; instead, \component{EA} lifts overall test MRR over \predictoronly{} on every dataset ($0.4099\to 0.4305$, $0.3082\to 0.3123$, $0.5177\to 0.5253$, and $0.1055\to 0.1084$).

When decomposed by role (\autoref{tab:role-cohort-results}), Stage II delivers balanced improvements across both unseen endpoints rather than trading one role against the other.
Specifically, relative to \predictoronly{}, Query-new MRR rises by $9.29\%$--$21.94\%$ (achieving the top score on all three ICEWS datasets), Answer-new MRR rises by $7.81\%$--$41.24\%$ (reaching $0.2176$ on GDELT, a $2.7\times$ improvement over TransFIR$^\ast$ and $6.4\times$ over AdaTKG$^\ast$), and Both-new MRR increases by $40.21\%$ on ICEWS14 and $68.06\%$ on ICEWS18 (on ICEWS05--15, Both-new comprises only $86$ queries; on GDELT, only $2$).
In addition, cohort $\operatorname{MRR}_{\mathrm{macro}}$ improves by $9.64\%$--$39.94\%$ over \predictoronly{} ($+16.93\%$ to $+150.54\%$ over the strongest baseline), hence confirming that gains generalize beyond high-frequency newcomers.

\begin{table*}[!t]
    \centering
    \small
    \caption{Stage I ablation prior to \component{EA} (bold: best per dataset and column). E, A, B, and C$_{\mathrm{M}}$ denote Emerging, Answer-new, Both-new, and cohort $\operatorname{MRR}_{\mathrm{macro}}$. On GDELT, \ablation{CSM} coincides with Full ($p_C=0$). Paired results after \component{EA} appear in the supplement.}
    \label{tab:predictor-components}
    \renewcommand{\arraystretch}{0.90}
    \begin{tabularx}{\textwidth}{lRRRRRRRR}
        \toprule
        \multicolumn{1}{c}{\multirow{2}{*}{Variant}}          & \multicolumn{4}{c}{ICEWS14}     & \multicolumn{4}{c}{ICEWS18}                                                                                                                 \\
        \cmidrule(lr){2-5}\cmidrule(l){6-9}
                                                              & E                               & A                           & B               & C$_{\mathrm{M}}$ & E               & A               & B                 & C$_{\mathrm{M}}$ \\
        \midrule
        \rowcolor{graywoea} Full predictor (\method{} w/o EA) & 0.1918                          & 0.1425                      & 0.1909          & 0.1027           & \textbf{0.1510} & 0.1025          & 0.1263            & 0.0705           \\
        w/o CSM                                               & 0.1928                          & 0.1475                      & 0.1748          & 0.1002           & 0.1496          & \textbf{0.1304} & \textbf{0.1320}   & \textbf{0.0848}  \\
        w/o QSM+CSM                                           & 0.1831                          & 0.1496                      & 0.1417          & 0.0965           & 0.1494          & 0.0861          & 0.0988            & 0.0606           \\
        w/o FCR+CSM                                           & 0.1936                          & 0.1527                      & 0.1831          & 0.1035           & 0.1459          & 0.1052          & 0.1173            & 0.0730           \\
        w/o FCR+QSM+CSM                                       & \textbf{0.2029}                 & \textbf{0.1658}             & \textbf{0.1937} & \textbf{0.1123}  & 0.1419          & 0.1104          & 0.0886            & 0.0725           \\
        w/o GMF+QSM+CSM                                       & 0.1625                          & 0.1213                      & 0.1489          & 0.0830           & 0.1062          & 0.0598          & 0.0847            & 0.0426           \\
        \midrule
        \multicolumn{1}{c}{\multirow{2}{*}{Variant}}          & \multicolumn{4}{c}{ICEWS05--15} & \multicolumn{4}{c}{GDELT}                                                                                                                   \\
        \cmidrule(lr){2-5}\cmidrule(l){6-9}
                                                              & E                               & A                           & B               & C$_{\mathrm{M}}$ & E               & A               & B                 & C$_{\mathrm{M}}$ \\
        \midrule
        \rowcolor{graywoea} Full predictor (\method{} w/o EA) & 0.2389                          & 0.1961                      & 0.1415          & 0.1442           & 0.1224          & 0.1541          & 0.000136          & 0.0909           \\
        w/o CSM                                               & \textbf{0.2438}                 & \textbf{0.2000}             & \textbf{0.1734} & 0.1459           & 0.1224          & 0.1541          & 0.000136          & 0.0909           \\
        w/o QSM+CSM                                           & 0.2384                          & 0.1967                      & 0.1534          & \textbf{0.1468}  & \textbf{0.1652} & \textbf{0.2137} & 0.000141          & \textbf{0.1262}  \\
        w/o FCR+CSM                                           & 0.2328                          & 0.1900                      & 0.0852          & 0.1407           & 0.1249          & 0.1284          & 0.000154          & 0.0762           \\
        w/o FCR+QSM+CSM                                       & 0.2378                          & 0.1851                      & 0.1425          & 0.1390           & 0.1535          & 0.1853          & 0.000161          & 0.1073           \\
        w/o GMF+QSM+CSM                                       & 0.2181                          & 0.1778                      & 0.1305          & 0.1326           & 0.0787          & 0.0816          & \textbf{0.000192} & 0.0524           \\
        \bottomrule
    \end{tabularx}
\end{table*}

\subsection{Stage I Ablation (RQ2)}
\label{sec:predictor-components}
To evaluate \textbf{RQ2}, \autoref{tab:predictor-components} ablates Stage I modules prior to \component{EA} against the complete Stage I predictor (\predictoronly{}, denoted Full predictor):
\begin{itemize}[leftmargin=*, nosep, topsep=1.5pt]
    \item \textbf{\ablation{CSM}}: disables catalog-side masking ($p_C=0$).
    \item \textbf{\ablation{QSM+CSM}}: disables both training masks ($p_H=p_C=0$).
    \item \textbf{\ablation{FCR+CSM}} and \textbf{\ablation{FCR+QSM+CSM}}: restrict memory readout to active query heads $\mathcal{U}_t^d$ with \component{QSM} enabled and disabled, respectively.
    \item \textbf{\ablation{GMF+QSM+CSM}}: zeroes $\mathbf{g}_e=\mathbf{0}$ with masks off, isolating the base pattern transfer backbone.
\end{itemize}
In addition, dedicated Stage II adapters are trained on every ablated predictor to examine cross-stage synergy (full paired tables appear in the supplementary material).

\noindent\textbf{Memory fusion and catalog readout.}
Comparing \ablation{QSM+CSM} against \ablation{GMF+QSM+CSM} isolates \component{GMF} at zero masking: gated catalog fusion boosts pre-\component{EA} Emerging MRR across all four datasets by $9.3\%$--$110.0\%$ ($+12.7\%$ on ICEWS14, $+40.7\%$ on ICEWS18, $+9.3\%$ on ICEWS05--15, and $+110.0\%$ on GDELT), with even larger gains on Answer-new ($+10.6\%$--$+162.0\%$) and cohort $\operatorname{MRR}_{\mathrm{macro}}$ ($+10.7\%$--$+141.0\%$).
Furthermore, this advantage persists after pairing each variant with its own Stage II adapter ($+9.1\%$--$+26.6\%$ paired Emerging MRR), hence demonstrating that Stage II cannot substitute for a Stage I representation calibrated across the catalog.
Similarly, comparing \ablation{QSM+CSM} with \ablation{FCR+QSM+CSM} shows that enabling \component{FCR} at zero masking strengthens Answer-new and cohort $\operatorname{MRR}_{\mathrm{macro}}$ on long-horizon ICEWS05--15 and dense GDELT ($+15.3\%$ Answer-new and $+17.6\%$ cohort $\operatorname{MRR}_{\mathrm{macro}}$ on GDELT), where non-query catalog candidates accumulate dense histories that otherwise skew full-catalog competition.

\noindent\textbf{Availability masking across stream regimes.}
Since \component{QSM} and \component{CSM} deliberately drop past entity memory during training to simulate zero-support inference, their standalone Stage I scores reflect an expected regularization trade-off that depends strongly on dataset length, catalog breadth, and density.
To illustrate, on broad ICEWS18 dominated by zero-support debuts ($23{,}033$ entities; $70.35\%$ zero-support debut share), masking prevents the fusion gate from overfitting mature actors.
Although pre-\component{EA} \ablation{CSM} shows slightly higher standalone Answer-new, pairing each predictor with Stage II reveals that full \component{CSM} delivers a substantial cross-stage advantage, surpassing paired \ablation{CSM} by $+13.2\%$ Emerging, $+23.6\%$ Query-new, $+31.0\%$ Both-new, and $+8.3\%$ cohort $\operatorname{MRR}_{\mathrm{macro}}$, attributed to \component{EA} refining a geometry already calibrated for absent history.
Likewise, on short, sparse ICEWS14 ($365$ snapshots), unmasked local readout (\ablation{FCR+QSM+CSM}) fits the compact training graph slightly tighter prior to \component{EA}; nevertheless, once paired with Stage II, complete \method{} surpasses paired \ablation{FCR+QSM+CSM} across Emerging, Query-new, Both-new ($+10.6\%$), and cohort $\operatorname{MRR}_{\mathrm{macro}}$, while \component{CSM} lifts paired Both-new by $+7.2\%$ and cohort $\operatorname{MRR}_{\mathrm{macro}}$ by $+2.4\%$.
On the contrary, on long-horizon ICEWS05--15 ($4{,}017$ snapshots, $83.32\%$ post-debut queries) and high-frequency GDELT ($828$ facts/snapshot, $0.166\%$ emerging share), established temporal dynamics dominate; consequently, lighter or disabled masking (\ablation{QSM+CSM}) preserves denser native patterns and achieves higher standalone and paired scores ($0.2695$ and $0.2002$ paired Emerging MRR), consistent with validation selecting $p_C=0$ on GDELT.
Therefore, availability masking acts as a regularizer sensitive to stream regimes that is most impactful on shorter streams or streams with higher debut ratios.

\subsection{Stage II Ablation (RQ3)}
\label{sec:adapter-components}
For \textbf{RQ3}, \autoref{tab:adapter-components} ablates Stage II \component{EA} components while holding the frozen Stage I predictor fixed per dataset under the debiased, aligned setting ($\alpha>0$, $\mathbf{b}_{\mathrm{train}}$, $\lambda_{\mathrm{align}}=0.2$):
\begin{itemize}[leftmargin=*, nosep, topsep=1.5pt]
    \item \textbf{\ablation{CEW}}: replaces $C_e$ with independent covariance of matched marginal variance $(1+\gamma)I=1.5I$.
    \item \textbf{\ablation{LRE}}: restricts queues to the last $14$ snapshots (retaining $K=32$ and lifetime support labels).
    \item \textbf{\ablation{CEW+LRE}}: removes both modules jointly ($2\times 2$ design).
    \item \textbf{\ablation{RBS}}: removes the role-balanced branch ($\mathcal{L}_{\mathrm{II}}=\mathcal{L}_{\mathrm{sup}}+\lambda_{\mathrm{align}}\mathcal{L}_{\mathrm{align}}$).
\end{itemize}

\begin{table*}[!t]
    \centering
    \small
    \caption{Stage II \component{EA} ablation on the frozen Stage I predictor (bold: best per dataset and column). Metrics follow \autoref{tab:predictor-components}; variants are defined in \autoref{sec:adapter-components}, with complete metric panels and alignment/shortcut controls in the supplement.}
    \label{tab:adapter-components}
    \renewcommand{\arraystretch}{0.90}
    \begin{tabularx}{\textwidth}{lRRRRRRRR}
        \toprule
        \multicolumn{1}{c}{\multirow{2}{*}{Variant}} & \multicolumn{4}{c}{ICEWS14}     & \multicolumn{4}{c}{ICEWS18}                                                                                                                 \\
        \cmidrule(lr){2-5}\cmidrule(l){6-9}
                                                     & E                               & A                           & B               & C$_{\mathrm{M}}$ & E               & A               & B                 & C$_{\mathrm{M}}$ \\
        \midrule
        \rowcolor{graywoea} \method{} w/o EA         & 0.1918                          & 0.1425                      & 0.1909          & 0.1027           & 0.1510          & 0.1025          & 0.1263            & 0.0705           \\
        \rowcolor{grayinspire} \method{}             & 0.2264                          & 0.1679                      & 0.2677          & 0.1228           & 0.1877          & 0.1291          & \textbf{0.2122}   & 0.0963           \\
        w/o CEW                                      & 0.2242                          & \textbf{0.1737}             & 0.2550          & \textbf{0.1264}  & \textbf{0.1950} & \textbf{0.1461} & 0.1961            & \textbf{0.1059}  \\
        w/o LRE                                      & 0.2219                          & 0.1636                      & 0.2471          & 0.1187           & 0.1800          & 0.1172          & 0.1634            & 0.0844           \\
        w/o CEW+LRE                                  & 0.2202                          & 0.1619                      & 0.2527          & 0.1184           & 0.1858          & 0.1243          & 0.1823            & 0.0898           \\
        w/o RBS                                      & \textbf{0.2267}                 & 0.1703                      & \textbf{0.2859} & 0.1256           & 0.1748          & 0.1262          & 0.1920            & 0.0881           \\
        \midrule
        \multicolumn{1}{c}{\multirow{2}{*}{Variant}} & \multicolumn{4}{c}{ICEWS05--15} & \multicolumn{4}{c}{GDELT}                                                                                                                   \\
        \cmidrule(lr){2-5}\cmidrule(l){6-9}
                                                     & E                               & A                           & B               & C$_{\mathrm{M}}$ & E               & A               & B                 & C$_{\mathrm{M}}$ \\
        \midrule
        \rowcolor{graywoea} \method{} w/o EA         & 0.2389                          & 0.1961                      & 0.1415          & 0.1442           & 0.1224          & 0.1541          & 0.000136          & 0.0909           \\
        \rowcolor{grayinspire} \method{}             & \textbf{0.2587}                 & \textbf{0.2114}             & 0.1094          & \textbf{0.1581}  & 0.1623          & 0.2176          & 0.000134          & 0.1272           \\
        w/o CEW                                      & 0.2560                          & 0.2106                      & 0.1100          & 0.1578           & \textbf{0.1928} & \textbf{0.2674} & 0.000144          & \textbf{0.1597}  \\
        w/o LRE                                      & 0.2563                          & 0.2070                      & 0.1054          & 0.1562           & 0.1744          & 0.2315          & 0.000153          & 0.1376           \\
        w/o CEW+LRE                                  & 0.2554                          & 0.2049                      & 0.1194          & 0.1538           & 0.1721          & 0.2313          & 0.000153          & 0.1361           \\
        w/o RBS                                      & 0.2516                          & 0.2015                      & \textbf{0.1537} & 0.1552           & 0.1538          & 0.2019          & \textbf{0.000157} & 0.1187           \\
        \bottomrule
    \end{tabularx}
\end{table*}

\noindent\textbf{Long-range evidence (\component{LRE}).}
Retaining sparse interactions over the full prefix lifts Emerging, Answer-new, Both-new, and cohort $\operatorname{MRR}_{\mathrm{macro}}$ across all three ICEWS datasets.
In particular, on sparse ICEWS18 with a broad catalog, \component{LRE} improves over \ablation{LRE} by $+4.3\%$ Emerging, $+10.2\%$ Answer-new, $+29.9\%$ Both-new, and $+14.1\%$ cohort $\operatorname{MRR}_{\mathrm{macro}}$, since emerging actors in geopolitical streams interact sporadically across long intervals and a $14$-snapshot window prematurely evicts informative early links.
On the contrary, on highly dense GDELT ($828$ facts/snapshot), rapid short-term news turnover causes older interactions to become less predictive of immediate links; hence, the $14$-snapshot window attains higher Emerging MRR ($+7.5\%$ over full prefix queues) at a slightly larger residual offset between zero-support and observed candidates, illustrating a temporal drift trade-off between sparse and high-frequency streams.

\noindent\textbf{Correlated evidence weighting (\component{CEW}).}
Comparing \method{} against \ablation{CEW} reveals a fundamental accuracy--calibration trade-off driven by bursty interaction redundancy (\textbf{C2}).
Without correlation discounting, independent aggregation ($1.5I$) excessively contracts posterior variance $v_{o'}$ when established entities exhibit repetitive bursts, consequently inflating the variance contrast between zero-support and observed candidates.
Although this uncalibrated contrast boosts Answer-new MRR in \ablation{CEW} on bursty ICEWS18 and GDELT, it nevertheless degrades overall test MRR on all four datasets ($0.4305$ vs.\ $0.4254$ on ICEWS14, $0.3123$ vs.\ $0.3105$ on ICEWS18, $0.5253$ vs.\ $0.5213$ on ICEWS05--15, and $0.1084$ vs.\ $0.1055$ on GDELT) while widening nontarget residual offsets.
In contrast, with \component{CEW}, \method{} maintains the strongest full-catalog calibration on every dataset while improving Emerging and Query-new MRR on ICEWS14 and ICEWS05--15 and lifting Both-new MRR by $+8.2\%$ on ICEWS18.

\noindent\textbf{Role-balanced supervision (\component{RBS}) and calibration verification.}
Removing \component{RBS} degrades Query-new MRR on every dataset (up to $-9.2\%$ on ICEWS18, where \component{RBS} lifts Emerging MRR by $+7.4\%$ and cohort $\operatorname{MRR}_{\mathrm{macro}}$ by $+9.3\%$) and reduces Emerging MRR on ICEWS05--15 ($+2.8\%$ gain from \component{RBS}) and GDELT ($+5.5\%$), confirming that the role-balanced co-answer branch prevents support-stratified cross-entropy from neglecting unseen query heads.
Furthermore, supplementary calibration controls verify that dropping gauge alignment ($\lambda_{\mathrm{align}}=0$) leaks a blanket zero-support residual offset back into test scores (hurting overall test MRR on ICEWS14 and GDELT), and that relative to a $1$-parameter zero-support shortcut control ($z_\theta(q,o)+\beta\mathbf{1}[n_o(t)=0]$), calibrated \method{} lifts Query-new MRR by $18.1\%$--$40.0\%$ across all four datasets and doubles uncovered Answer-new MRR on ICEWS18 ($0.0131$ vs.\ $0.0065$), confirming genuine semantic and structural discrimination.

\begin{figure*}[!t]
    \centering
    \includegraphics[width=0.84\textwidth]{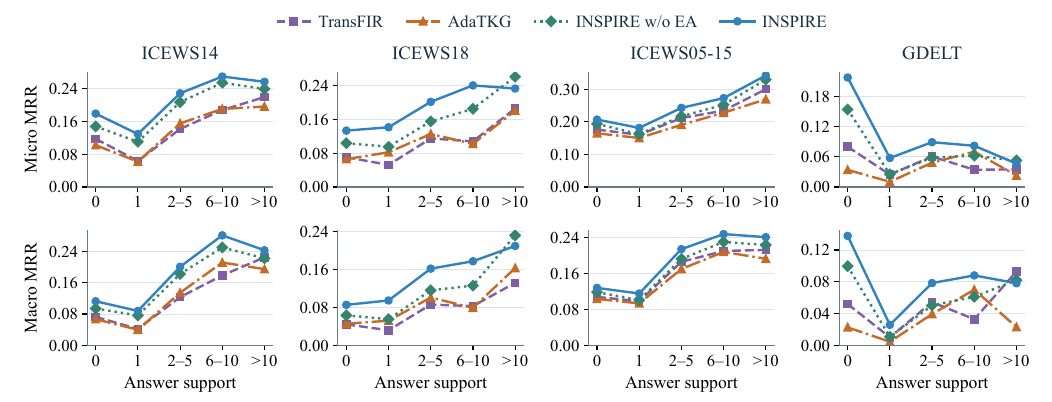}
    \caption{Emerging-answer cohort $\operatorname{MRR}_{\mathrm{micro}}$ (top) and $\operatorname{MRR}_{\mathrm{macro}}$ (bottom) across answer support strata ($0$ to $>10$). Cohort queries/identities: 2,200/1,040 (ICEWS14), 5,113/2,692 (ICEWS18), 12,249/1,593 (ICEWS05--15), and 2,080/785 (GDELT).}
    \Description{Four datasets and five support strata comparing TransFIR, AdaTKG, INSPIRE w/o EA, and INSPIRE on micro and macro MRR.}
    \label{fig:cohort-support}
\end{figure*}

\subsection{Longitudinal Cohort Analysis (RQ4)}
\label{sec:cohort-results}
For \textbf{RQ4}, \autoref{fig:cohort-support} tracks the emerging-answer cohort $\mathcal{C}_S$ across natural answer support strata ($0, 1, 2\text{--}5, 6\text{--}10, >10$).

Across ICEWS14 and ICEWS05--15, \method{} stays strictly above TransFIR$^\ast$, AdaTKG$^\ast$, and \predictoronly{} in every support bin from zero-shot debut ($0$) to mature accumulation ($>10$) on both $\operatorname{MRR}_{\mathrm{micro}}$ and $\operatorname{MRR}_{\mathrm{macro}}$.
Specifically, on long-horizon ICEWS05--15 ($83.32\%$ post-debut cohort queries), \component{EA} sustains positive gains even at $>10$ support ($6{,}329$ queries, $198$ identities), outperforming TransFIR$^\ast$ by $+14.1\%$ in $\operatorname{MRR}_{\mathrm{micro}}$ ($0.3427$ vs.\ $0.3004$) and $+13.3\%$ in $\operatorname{MRR}_{\mathrm{macro}}$ ($0.2404$ vs.\ $0.2121$).
Similarly, on ICEWS14 at $2$--$5$ support ($330$ queries, $143$ identities), \component{EA} adds $+10.5\%$ $\operatorname{MRR}_{\mathrm{micro}}$ and $+10.0\%$ $\operatorname{MRR}_{\mathrm{macro}}$ over \predictoronly{}.
Furthermore, on ICEWS18 and GDELT, \method{} leads all comparators throughout the entire zero-shot and few-shot range ($0$ through $6$--$10$): on ICEWS18 at $6$--$10$ support, \component{EA} lifts $\operatorname{MRR}_{\mathrm{micro}}$ by $+30.0\%$ ($0.1852\to 0.2407$) and $\operatorname{MRR}_{\mathrm{macro}}$ by $+40.8\%$ ($0.1263\to 0.1778$), whereas on GDELT at support $1$, \component{EA} more than doubles both $\operatorname{MRR}_{\mathrm{micro}}$ ($+128.6\%$) and $\operatorname{MRR}_{\mathrm{macro}}$ ($+138.9\%$).
Nevertheless, only in the small $>10$ tail of ICEWS18 ($16$ identities) and GDELT ($9$ identities) does \method{} dip modestly below \predictoronly{}, while remaining well above both temporal inductive baselines in $\operatorname{MRR}_{\mathrm{micro}}$.

\begin{figure*}[!t]
    \centering
    \includegraphics[width=0.86\textwidth]{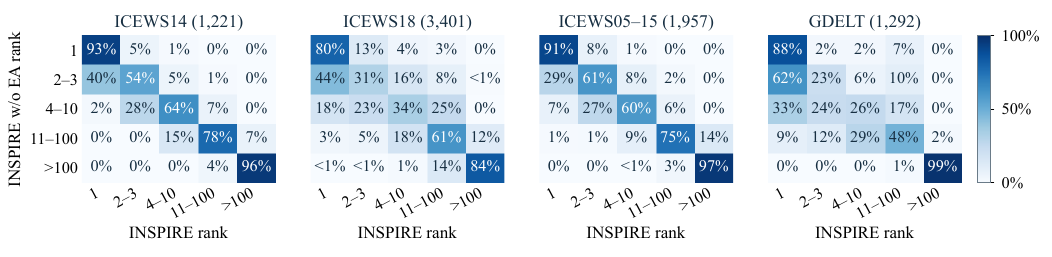}
    \caption{Answer-new rank band transition matrices from \predictoronly{} (rows) to \method{} (columns); rows sum to $100\%$, and panel titles report Answer-new query counts. Below-diagonal cells indicate rank promotions.}
    \Description{Four 5x5 heatmaps showing Answer-new rank transitions between INSPIRE w/o EA and INSPIRE across ICEWS14, ICEWS18, ICEWS05-15, and GDELT.}
    \label{fig:answer-rank-transitions}
\end{figure*}

\subsection{Mechanism Analysis (RQ5)}
\label{sec:evidence-mechanisms}
\begin{figure}[!t]
    \centering
    \includegraphics[width=0.90\linewidth]{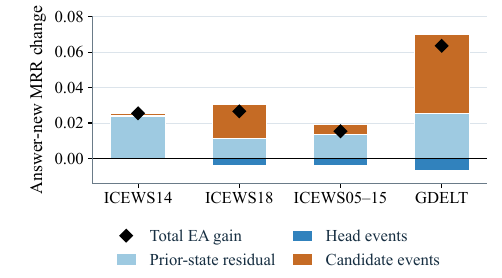}
    \caption{Shapley decomposition of Answer-new MRR gains at fixed checkpoints $(\theta^\ast,\phi^\ast)$ into the trained semantic prior residual (\component{SP}) and head vs.\ candidate event updates.}
    \Description{Single-column stacked bar chart decomposing Answer-new MRR gains into prior residual, head events, and candidate events across four datasets.}
    \label{fig:evidence-mechanisms}
\end{figure}
\autoref{tab:role-cohort-results} raises a central question (\textbf{RQ5}): without prior shortcuts induced by balanced losses, how does \component{EA} boost Answer-new MRR by $7.81\%$--$41.24\%$ when target $o$ has zero prior events ($k_o=0$)?
To answer this question, we freeze trained checkpoints $(\theta^\ast,\phi^\ast)$, run seven inference switches (predictor only, prior residual only, full evidence, head events only, candidate events only, mean updates only, and variance updates only), and subsequently apply exact $2\times 2$ Shapley allocations to partition the net event gain ($\mathrm{Full}-\mathrm{Prior}$) between head and candidate events (\autoref{fig:evidence-mechanisms}).

\noindent\textbf{Semantic prior and candidate score competition.}
\autoref{fig:evidence-mechanisms} shows that under $\alpha>0$, both the learned Prior residual and Candidate events yield strictly positive Answer-new gains on every dataset, whereas Head events stay nearly zero or slightly negative.
Specifically, on sparse ICEWS14 and long-horizon ICEWS05--15, textual priors contribute the largest share alongside positive candidate reinforcement; in contrast, on ICEWS18 with a broad catalog and dense GDELT, Candidate events drive $72.8\%$ and $69.9\%$ of the total gain.
While unseen target $o$ remains at its prior $(\boldsymbol{\mu}_{0,o},\lambda_{0,o}^{-1})$, correlated evidence updates competing established candidates $o'\neq o$, contracting $v_{o'}$ and sharpening the Mahalanobis penalty $d_{h,o'}/V_{h,o'}$ (\autoref{eq:final-score}) to demote mismatched distractors.
Furthermore, supplementary mean/variance decompositions across both roles confirm the coupled Gaussian effect: posterior variance contraction is strictly positive on Answer-new across all four datasets, whereas posterior mean updates act constructively on Query-new across ICEWS and on Answer-new in GDELT.

\noindent\textbf{Selective promotion of moderately ranked targets.}
In addition, \autoref{fig:answer-rank-transitions} confirms the aforementioned mechanism across the Answer-new rank distribution: \component{EA} preserves $80\%$--$93\%$ of Rank-1 predictions, promotes $29\%$--$62\%$ of Rank $2$--$3$ targets to Rank 1, lifts $30\%$--$57\%$ of Rank $4$--$10$ targets into Top-3, and moves $11\%$--$50\%$ of Rank $11$--$100$ targets into Top-10.
Meanwhile, $84\%$--$99\%$ of low-ranked ($>100$) targets remain $>100$ ($0\%$ jumping to Rank 1 on ICEWS14, ICEWS05--15, and GDELT; $<1\%$ on ICEWS18), hence confirming selective competition refinement rather than blanket zero-support inflation.

\subsection{Sensitivity and Efficiency (RQ6)}
\label{sec:sensitivity-cost}
\begin{figure*}[!t]
    \centering
    \includegraphics[width=0.86\textwidth]{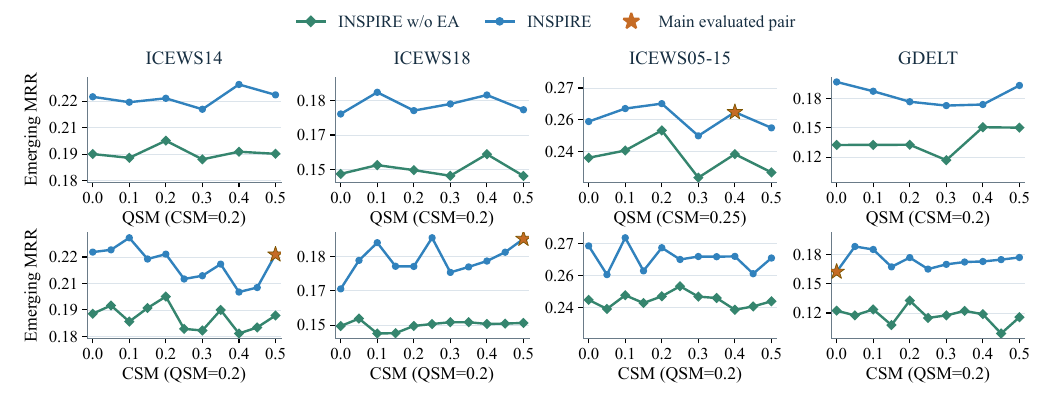}
    \caption{Test Emerging MRR sensitivity across orthogonal 1D \component{QSM} ($p_H$) and \component{CSM} ($p_C$) slices (fixed companion rates in axis labels). Stars mark main evaluated pairs, each lying on one slice per dataset (bottom row on ICEWS14/ICEWS18/GDELT; top row on ICEWS05--15).}
    \Description{Eight panels showing predictor and complete model Emerging MRR across QSM and CSM masking sweeps with main evaluated configuration stars.}
    \label{fig:masking-sensitivity}
\end{figure*}

To answer \textbf{RQ6}, \autoref{fig:masking-sensitivity} sweeps \component{QSM} ($p_H\in[0,0.5]$, step $0.1$) and \component{CSM} ($p_C\in[0,0.5]$, step $0.05$) along orthogonal 1D slices.
Across all $64$ evaluated $(p_H,p_C)$ pairs ($16$ per dataset), debiased \component{EA} beats its paired Stage I predictor in Emerging, Answer-new, Cohort $\operatorname{MRR}_{\mathrm{micro}}$, and Cohort $\operatorname{MRR}_{\mathrm{macro}}$ at every point ($64/64$).
Moreover, fixing Stage I predictors selected on validation leaves our reported GDELT Emerging MRR ($0.1623$) conservative relative to neighboring sweep points ($0.1650$--$0.1971$).

Furthermore, Stage II hyperparameters are equally stable: on ICEWS14 and ICEWS18 ($p_H=p_C=0.2$), validation Emerging MRR shifts by $<0.0008$ between $K=32$ and $K=64$ ($<0.0013$ across $K\in\{16,32,64\}$) and stays within $0.004$ across $\gamma\in\{0,0.1,0.5,1,2\}$ (positive $\gamma$ consistently lifting validation Emerging MRR on ICEWS14 and Query-new MRR on ICEWS18 over $\gamma=0$).
In addition, \autoref{tab:efficiency} shows that freezing $\theta$ and caching Cholesky sums (\autoref{sec:complexity}) keeps Stage II peak GPU memory to $2.17$--$8.24$\,GiB and training time to $23\%$--$41\%$ of Stage I on ICEWS ($62\%$ on GDELT); extended latencies appear in the supplement.

\begin{table}[!t]
    \centering\small
    \caption{Standalone training time (with validation and checkpointing) and peak Stage II GPU memory at $p_H=p_C=0.2$.}
    \label{tab:efficiency}
    \setlength{\tabcolsep}{3pt}
    \begin{tabular*}{\linewidth}{@{\extracolsep{\fill}}lrrr@{}}
        \toprule
        Dataset & Stage I (s) & Stage II (s) & GPU (GiB) \\
        \midrule
        ICEWS14 & 2,931.1 & 683.5 & 2.50 \\
        ICEWS18 & 12,359.7 & 5,018.0 & 7.14 \\
        ICEWS05--15 & 20,772.7 & 5,716.5 & 2.17 \\
        GDELT & 18,882.8 & 11,619.8 & 8.24 \\
        \bottomrule
    \end{tabular*}
\end{table}

\noindent\textbf{Paradigm synthesis.}
Therefore, \textbf{RQ1}--\textbf{RQ6} distill two general principles: (i) capacity--sparsity decoupling freezes a shared backbone calibrated for history availability while fitting a compact correlation-weighted residual to avoid sparse overfitting, and (ii) catalog-wide competition matters as much as newcomer encoding, since accurately calibrating established distractors enables emerging actors to surface reliably from debut through sparse growth.

\section{Related Work}
\label{sec:related_work}

\noindent\textbf{Temporal history and event forecasting.}
Early dynamic encoders modeled continuous TKG evolution via temporal point processes (Know-Evolve~\cite{trivedi2017knowevolve}), recurrent neighborhood aggregation (RE-Net~\cite{jin2020renet}), and neural differential equations (TANGO~\cite{han2021tango}).
Subsequently, snapshot models refined historical reuse through copy--generation modes (CyGNet~\cite{zhu2021cygnet}), relation-aware graph convolutions (RE-GCN~\cite{li2021regcn}), and local--global repetition (TiRGN~\cite{li2022tirgn}).
Furthermore, recent works model query--candidate subgraphs (HiSMatch~\cite{li2022hismatch}), multigranularity hypergraphs (MGESL~\cite{shi2024mgesl}), local--global contrastive learning (LogCL~\cite{chen2024logcl}), concurrent and evolutionary multigraphs (LMS~\cite{zhang2024multigraph}), and query relevance structuring (HisRES~\cite{zhang2025hisres}).
However, since these transductive models depend on entity lookup tables and dense histories, accuracy degrades sharply for emerging entities.

\noindent\textbf{Structural, logical, and language model reasoning.}
Symbolic and structural approaches extract interpretable evidence via temporal subgraphs and logical rules (xERTE~\cite{han2021xerte}, TLogic~\cite{liu2022tlogic}, TILP~\cite{xiong2023tilp}), evolutionary event chain Transformers (ECEformer~\cite{fang2024eceformer}), and hybrid inductive--deductive--abductive hypergraph reasoning~\cite{tian2026hybrid}.
In addition, language model approaches serialize facts into time-aware or in-context prompts (PPT~\cite{xu2023ppt}, ICL~\cite{lee2023icl}), generative decoding guided by retrieved rules (GenTKG~\cite{liao2024gentkg}), and topology-guided LLM logical learning~\cite{gao2026structural}.
On the contrary, \method{} performs full-catalog discriminative ranking that unites textual priors with correlation-adjusted statistics over long-range prefixes.

\noindent\textbf{Inductive transfer and adaptive entities.}
Static inductive methods generalize beyond fixed vocabularies via relational compositions or meta-knowledge (CompGCN~\cite{vashishth2020compgcn}, MorsE~\cite{chen2022morse}, InGram~\cite{lee2023ingram}, ULTRA~\cite{galkin2024ultra}), whereas distributed and federated graph systems study collaborative alignment over decentralized dynamic graphs~\cite{islam2026prism,zhang2025memshare,zheng2020privacypreserved,zheng2022distributed,zheng2023fair,wang2024fedgcl,zheng2025metacomputing}.
In temporal settings, OntoTKGE~\cite{lin2026ontotkge} transfers concept dynamics via ontologies, TransFIR~\cite{zhao2026transfir} maps unseen entities to behavioral codebooks via textual clusters, and AdaTKG~\cite{lee2026adatkg} maintains online per-entity EMA memory.
Nevertheless, TransFIR omits catalog-wide dynamic memory and AdaTKG's updates centered on active queries degrade Answer-new ranking (\autoref{tab:role-cohort-results}); hence, \method{} resolves this gap in Stage I via \component{GMF}, \component{FCR}, \component{QSM}, and \component{CSM}.

\noindent\textbf{Few-shot adaptation and correlated evidence.}
Few-shot TKG models target scarce observations through meta-learned neighbor aggregation (MetaTKGR~\cite{wang2022metatkgr}) or mutual distillation (MLEMKD~\cite{qian2025mlemkd}).
Furthermore, freezing pretrained backbones while tuning compact adapters~\cite{houlsby2019adapters} or retrieval caches~\cite{zhang2022tipadapter} supports data-efficient specialization, whereas Set Transformer~\cite{lee2019settransformer} and Bayesian Context Aggregation~\cite{volpp2021bayesian} aggregate variable-size sets via attention or independent Gaussian updates.
Accordingly, Stage II (\component{EA}) adapts this Bayesian updating paradigm over a frozen backbone to full-catalog TKG ranking with closed-form \component{CEW} over bounded \component{LRE} queues under debiased \component{RBS} and \component{CSS}.

\section{Conclusion}
\label{sec:conclusion}

Forecasting links for emerging entities in TKGs requires reasoning across a skewed continuum from zero-support debuts to sparse, bursty trajectories.
To address the aforementioned challenge, we presented \method{}, a two-stage framework decoupling availability-aware shared patterns (\component{GMF}, \component{FCR}, \component{QSM}, \component{CSM}) from lightweight evidence adaptation (\component{EA}) with textual priors, long-range queues, Gram matrix correlation weighting, and debiased, aligned balanced supervision.

Across all four benchmarks, \method{} improves Emerging MRR by $16.73\%$--$88.83\%$ over the strongest baselines, resolves adaptive memory Answer-new degradation, lifts overall test MRR on every dataset, and sustains cohort $\operatorname{MRR}_{\mathrm{macro}}$ gains across support growth.
Furthermore, decompositions and rank transition analyses at fixed checkpoints attribute calibrated Answer-new gains to semantic priors and candidate score competition, demoting mismatched established distractors and promoting plausible moderately ranked newcomers without zero-support shortcuts.

Finally, beyond preregistered catalogs and static names, promising future directions include open-vocabulary entity discovery under alias drift, evolving multimodal priors or priors grounded in documents, and selective forgetting adaptive to regime shifts with asynchronous inference on individual queries.\label{sec:body-end}

\FloatBarrier

\bibliographystyle{ACM-Reference-Format}
\bibliography{references}

\clearpage
\appendix
\setlength{\dbltextfloatsep}{5pt plus 1.5pt minus 1.5pt}
\setlength{\dblfloatsep}{4pt plus 1.5pt minus 1.5pt}
\setlength{\textfloatsep}{5pt plus 1.5pt minus 1.5pt}
\setlength{\abovecaptionskip}{2.5pt}
\setlength{\belowcaptionskip}{1.0pt}
\renewcommand{\topfraction}{0.95}
\renewcommand{\bottomfraction}{0.92}
\renewcommand{\dbltopfraction}{0.95}
\renewcommand{\textfraction}{0.05}
\renewcommand{\dblfloatpagefraction}{0.85}
\setcounter{table}{0}
\renewcommand{\thetable}{S\arabic{table}}
\setcounter{figure}{0}
\renewcommand{\thefigure}{S\arabic{figure}}
\makeatletter
\renewcommand{\theHtable}{S\arabic{table}}
\renewcommand{\theHfigure}{S\arabic{figure}}
\makeatother
\twocolumn[\begin{center}\LARGE\bfseries Supplemental Material\end{center}\vspace{8pt}]
\section{Evaluation Units and Component Definitions}
These supplemental tables and figures accompany the main manuscript and report complete evaluations across all six metrics under fixed random seed $42$. Specifically, E/Q/A/B denote Emerging, Query-new, Answer-new, and Both-new MRR, while C$_\mu$/C$_{\mathrm{M}}$ denote emerging-answer cohort $\operatorname{MRR}_{\mathrm{micro}}$ and $\operatorname{MRR}_{\mathrm{macro}}$. GDELT Both-new comprises two directed queries. The standalone predictor evaluation (\autoref{tab:supp-predictor}) and paired adaptation evaluation (\autoref{tab:supp-paired}) are conducted separately, whereas the Stage II adapter controls (\autoref{tab:supp-adapter}) share the frozen Stage I predictor; known catalog, directional batch, and history construction protocols match the main paper.
\begin{table*}[!b]
    \centering
    \small
    \renewcommand{\arraystretch}{0.98}
    \caption{Stage I predictor ablations before \component{EA}: all six focal MRR metrics (E/Q/A/B: Emerging/Query-new/Answer-new/Both-new; C$_\mu$/C$_{\mathrm{M}}$: cohort $\operatorname{MRR}_{\mathrm{micro}}$/$\operatorname{MRR}_{\mathrm{macro}}$). GDELT w/o CSM coincides with the reference ($p_C=0$).}
    \label{tab:supp-predictor}
    \begin{tabular*}{\textwidth}{@{\extracolsep{\fill}}clrrrrrr@{}}
        \toprule
        \multicolumn{1}{c}{Dataset} & \multicolumn{1}{c}{Variant} & E & Q & A & B & C$_\mu$ & C$_{\mathrm{M}}$ \\
        \midrule
        \multirow{6}{*}{ICEWS14} & \cellcolor{graywoea}Full predictor (\method{} w/o EA) & \cellcolor{graywoea}0.1918 & \cellcolor{graywoea}0.2413 & \cellcolor{graywoea}0.1425 & \cellcolor{graywoea}0.1909 & \cellcolor{graywoea}0.1646 & \cellcolor{graywoea}0.1027 \\
        & w/o CSM & 0.1928 & 0.2404 & 0.1475 & 0.1748 & 0.1663 & 0.1002 \\
        & w/o QSM+CSM & 0.1831 & 0.2218 & 0.1496 & 0.1417 & 0.1637 & 0.0965 \\
        & w/o FCR+CSM & 0.1936 & 0.2358 & 0.1527 & 0.1831 & 0.1650 & 0.1035 \\
        & w/o FCR+QSM+CSM & 0.2029 & 0.2411 & 0.1658 & 0.1937 & 0.1682 & 0.1123 \\
        & w/o GMF+QSM+CSM & 0.1625 & 0.2054 & 0.1213 & 0.1489 & 0.1330 & 0.0830 \\
        \midrule
        \multirow{6}{*}{ICEWS18} & \cellcolor{graywoea}Full predictor (\method{} w/o EA) & \cellcolor{graywoea}0.1510 & \cellcolor{graywoea}0.2009 & \cellcolor{graywoea}0.1025 & \cellcolor{graywoea}0.1263 & \cellcolor{graywoea}0.1185 & \cellcolor{graywoea}0.0705 \\
        & w/o CSM & 0.1496 & 0.1698 & 0.1304 & 0.1320 & 0.1382 & 0.0848 \\
        & w/o QSM+CSM & 0.1494 & 0.2156 & 0.0861 & 0.0988 & 0.1072 & 0.0606 \\
        & w/o FCR+CSM & 0.1459 & 0.1883 & 0.1052 & 0.1173 & 0.1193 & 0.0730 \\
        & w/o FCR+QSM+CSM & 0.1419 & 0.1765 & 0.1104 & 0.0886 & 0.1211 & 0.0725 \\
        & w/o GMF+QSM+CSM & 0.1062 & 0.1539 & 0.0598 & 0.0847 & 0.0730 & 0.0426 \\
        \midrule
        \multirow{6}{*}{ICEWS05--15} & \cellcolor{graywoea}Full predictor (\method{} w/o EA) & \cellcolor{graywoea}0.2389 & \cellcolor{graywoea}0.2859 & \cellcolor{graywoea}0.1961 & \cellcolor{graywoea}0.1415 & \cellcolor{graywoea}0.2718 & \cellcolor{graywoea}0.1442 \\
        & w/o CSM & 0.2438 & 0.2906 & 0.2000 & 0.1734 & 0.2754 & 0.1459 \\
        & w/o QSM+CSM & 0.2384 & 0.2839 & 0.1967 & 0.1534 & 0.2817 & 0.1468 \\
        & w/o FCR+CSM & 0.2328 & 0.2820 & 0.1900 & 0.0852 & 0.2580 & 0.1407 \\
        & w/o FCR+QSM+CSM & 0.2378 & 0.2947 & 0.1851 & 0.1425 & 0.2569 & 0.1390 \\
        & w/o GMF+QSM+CSM & 0.2181 & 0.2623 & 0.1778 & 0.1305 & 0.2414 & 0.1326 \\
        \midrule
        \multirow{6}{*}{GDELT} & \cellcolor{graywoea}Full predictor (\method{} w/o EA) & \cellcolor{graywoea}0.1224 & \cellcolor{graywoea}0.0909 & \cellcolor{graywoea}0.1541 & \cellcolor{graywoea}0.000136 & \cellcolor{graywoea}0.1149 & \cellcolor{graywoea}0.0909 \\
        & w/o CSM & 0.1224 & 0.0909 & 0.1541 & 0.000136 & 0.1149 & 0.0909 \\
        & w/o QSM+CSM & 0.1652 & 0.1169 & 0.2137 & 0.000141 & 0.1586 & 0.1262 \\
        & w/o FCR+CSM & 0.1249 & 0.1215 & 0.1284 & 0.000154 & 0.0947 & 0.0762 \\
        & w/o FCR+QSM+CSM & 0.1535 & 0.1218 & 0.1853 & 0.000161 & 0.1337 & 0.1073 \\
        & w/o GMF+QSM+CSM & 0.0787 & 0.0759 & 0.0816 & 0.000192 & 0.0693 & 0.0524 \\
        \bottomrule
    \end{tabular*}
\end{table*}

\newpage
For Stage I, we ablate catalog-side masking (\ablation{CSM}) and both masking channels (\ablation{QSM+CSM}). We further remove full-catalog readout (\ablation{FCR+CSM} and \ablation{FCR+QSM+CSM}) as well as gated memory fusion (\ablation{GMF+QSM+CSM}). Stage II controls share the identical frozen Stage I predictor per dataset under strictly positive $\alpha$, training-only prior debiasing across support buckets, and global residual alignment $\lambda_{\mathrm{align}}=0.2$. Specifically, \ablation{CEW} uses independent covariance with matched marginal variance $1.5I$, \ablation{LRE} limits queues to the preceding $14$ snapshots while retaining lifetime support labels, \ablation{CEW+LRE} combines both changes, and \ablation{RBS} drops the role-balanced branch. The italicized diagnostic control \textit{w/o Align} sets $\lambda_{\mathrm{align}}=0$ to illustrate unregularized zero-support gauge leakage, inflating covered nontarget gaps between zero-support and observed candidates to $+1.350$, $+0.672$, and $+1.186$ logits on ICEWS14, ICEWS18, and ICEWS05--15 and reducing overall test MRR on ICEWS14 and GDELT.
\clearpage

\begin{table*}[!tp]
    \centering
    \small
    \renewcommand{\arraystretch}{0.94}
    \caption{Stage I predictor ablations paired with dedicated Stage II \component{EA} adapters across all six focal MRR metrics (definitions follow \autoref{tab:supp-predictor}). Variant names identify the Stage I modification; GDELT w/o CSM coincides with the reference ($p_C=0$).}
    \label{tab:supp-paired}
    \begin{tabular*}{\textwidth}{@{\extracolsep{\fill}}clrrrrrr@{}}
        \toprule
        \multicolumn{1}{c}{Dataset} & \multicolumn{1}{c}{Variant} & E & Q & A & B & C$_\mu$ & C$_{\mathrm{M}}$ \\
        \midrule
        \multirow{6}{*}{ICEWS14} & \cellcolor{grayinspire}\method{} & \cellcolor{grayinspire}0.2264 & \cellcolor{grayinspire}0.2797 & \cellcolor{grayinspire}0.1679 & \cellcolor{grayinspire}0.2677 & \cellcolor{grayinspire}0.1913 & \cellcolor{grayinspire}0.1228 \\
        & w/o CSM & 0.2278 & 0.2817 & 0.1711 & 0.2499 & 0.1936 & 0.1200 \\
        & w/o QSM+CSM & 0.2143 & 0.2632 & 0.1654 & 0.2151 & 0.1839 & 0.1128 \\
        & w/o FCR+CSM & 0.2280 & 0.2785 & 0.1739 & 0.2570 & 0.1930 & 0.1232 \\
        & w/o FCR+QSM+CSM & 0.2234 & 0.2741 & 0.1704 & 0.2420 & 0.1856 & 0.1219 \\
        & w/o GMF+QSM+CSM & 0.1965 & 0.2372 & 0.1505 & 0.2384 & 0.1639 & 0.1076 \\
        \midrule
        \multirow{6}{*}{ICEWS18} & \cellcolor{grayinspire}\method{} & \cellcolor{grayinspire}0.1877 & \cellcolor{grayinspire}0.2449 & \cellcolor{grayinspire}0.1291 & \cellcolor{grayinspire}0.2122 & \cellcolor{grayinspire}0.1501 & \cellcolor{grayinspire}0.0963 \\
        & w/o CSM & 0.1658 & 0.1982 & 0.1337 & 0.1621 & 0.1453 & 0.0890 \\
        & w/o QSM+CSM & 0.1725 & 0.2337 & 0.1114 & 0.1707 & 0.1401 & 0.0823 \\
        & w/o FCR+CSM & 0.1719 & 0.2132 & 0.1322 & 0.1430 & 0.1461 & 0.0913 \\
        & w/o FCR+QSM+CSM & 0.1646 & 0.2123 & 0.1176 & 0.1515 & 0.1368 & 0.0846 \\
        & w/o GMF+QSM+CSM & 0.1363 & 0.1981 & 0.0747 & 0.1332 & 0.0974 & 0.0584 \\
        \midrule
        \multirow{6}{*}{ICEWS05--15} & \cellcolor{grayinspire}\method{} & \cellcolor{grayinspire}0.2587 & \cellcolor{grayinspire}0.3125 & \cellcolor{grayinspire}0.2114 & \cellcolor{grayinspire}0.1094 & \cellcolor{grayinspire}0.2878 & \cellcolor{grayinspire}0.1581 \\
        & w/o CSM & 0.2671 & 0.3181 & 0.2205 & 0.1682 & 0.2926 & 0.1646 \\
        & w/o QSM+CSM & 0.2695 & 0.3221 & 0.2206 & 0.1852 & 0.3044 & 0.1692 \\
        & w/o FCR+CSM & 0.2583 & 0.3077 & 0.2158 & 0.0991 & 0.2839 & 0.1611 \\
        & w/o FCR+QSM+CSM & 0.2549 & 0.3116 & 0.2026 & 0.1542 & 0.2887 & 0.1563 \\
        & w/o GMF+QSM+CSM & 0.2466 & 0.2921 & 0.2062 & 0.1320 & 0.2763 & 0.1550 \\
        \midrule
        \multirow{6}{*}{GDELT} & \cellcolor{grayinspire}\method{} & \cellcolor{grayinspire}0.1623 & \cellcolor{grayinspire}0.1072 & \cellcolor{grayinspire}0.2176 & \cellcolor{grayinspire}0.000134 & \cellcolor{grayinspire}0.1639 & \cellcolor{grayinspire}0.1272 \\
        & w/o CSM & 0.1623 & 0.1072 & 0.2176 & 0.000134 & 0.1639 & 0.1272 \\
        & w/o QSM+CSM & 0.2002 & 0.1308 & 0.2699 & 0.000146 & 0.2056 & 0.1611 \\
        & w/o FCR+CSM & 0.1955 & 0.1328 & 0.2584 & 0.000143 & 0.1926 & 0.1487 \\
        & w/o FCR+QSM+CSM & 0.1998 & 0.1381 & 0.2619 & 0.000149 & 0.2027 & 0.1534 \\
        & w/o GMF+QSM+CSM & 0.1603 & 0.1182 & 0.2026 & 0.000182 & 0.1645 & 0.1260 \\
        \bottomrule
    \end{tabular*}
\end{table*}

\begin{table*}[!tp]
    \centering
    \small
    \renewcommand{\arraystretch}{0.94}
    \caption{Stage II adapter controls sharing the identical frozen Stage I predictor across all six focal metrics (definitions follow \autoref{tab:supp-predictor}). Calibrated controls fix $\lambda_{\mathrm{align}}=0.2$; $^\ddagger$\textit{w/o Align} sets $\lambda_{\mathrm{align}}=0$, italicized for zero-support gauge leakage.}
    \label{tab:supp-adapter}
    \begin{tabular*}{\textwidth}{@{\extracolsep{\fill}}clrrrrrr@{}}
        \toprule
        \multicolumn{1}{c}{Dataset} & \multicolumn{1}{c}{Variant} & E & Q & A & B & C$_\mu$ & C$_{\mathrm{M}}$ \\
        \midrule
        \multirow{6}{*}{ICEWS14} & \cellcolor{grayinspire}\method{} & \cellcolor{grayinspire}0.2264 & \cellcolor{grayinspire}0.2797 & \cellcolor{grayinspire}0.1679 & \cellcolor{grayinspire}0.2677 & \cellcolor{grayinspire}0.1913 & \cellcolor{grayinspire}0.1228 \\
        & w/o CEW & 0.2242 & 0.2708 & 0.1737 & 0.2550 & 0.1953 & 0.1264 \\
        & w/o LRE & 0.2219 & 0.2770 & 0.1636 & 0.2471 & 0.1870 & 0.1187 \\
        & w/o CEW+LRE & 0.2202 & 0.2744 & 0.1619 & 0.2527 & 0.1875 & 0.1184 \\
        & w/o RBS & 0.2267 & 0.2757 & 0.1703 & 0.2859 & 0.1951 & 0.1256 \\
        & \textit{w/o Align}$^\ddagger$ & \textit{0.2520} & \textit{0.2832} & \textit{0.2127} & \textit{0.3153} & \textit{0.2271} & \textit{0.1512} \\
        \midrule
        \multirow{6}{*}{ICEWS18} & \cellcolor{grayinspire}\method{} & \cellcolor{grayinspire}0.1877 & \cellcolor{grayinspire}0.2449 & \cellcolor{grayinspire}0.1291 & \cellcolor{grayinspire}0.2122 & \cellcolor{grayinspire}0.1501 & \cellcolor{grayinspire}0.0963 \\
        & w/o CEW & 0.1950 & 0.2437 & 0.1461 & 0.1961 & 0.1674 & 0.1059 \\
        & w/o LRE & 0.1800 & 0.2438 & 0.1172 & 0.1634 & 0.1388 & 0.0844 \\
        & w/o CEW+LRE & 0.1858 & 0.2474 & 0.1243 & 0.1823 & 0.1451 & 0.0898 \\
        & w/o RBS & 0.1748 & 0.2225 & 0.1262 & 0.1920 & 0.1446 & 0.0881 \\
        & \textit{w/o Align}$^\ddagger$ & \textit{0.1996} & \textit{0.2470} & \textit{0.1499} & \textit{0.2392} & \textit{0.1719} & \textit{0.1129} \\
        \midrule
        \multirow{6}{*}{ICEWS05--15} & \cellcolor{grayinspire}\method{} & \cellcolor{grayinspire}0.2587 & \cellcolor{grayinspire}0.3125 & \cellcolor{grayinspire}0.2114 & \cellcolor{grayinspire}0.1094 & \cellcolor{grayinspire}0.2878 & \cellcolor{grayinspire}0.1581 \\
        & w/o CEW & 0.2560 & 0.3078 & 0.2106 & 0.1100 & 0.2870 & 0.1578 \\
        & w/o LRE & 0.2563 & 0.3122 & 0.2070 & 0.1054 & 0.2956 & 0.1562 \\
        & w/o CEW+LRE & 0.2554 & 0.3118 & 0.2049 & 0.1194 & 0.2907 & 0.1538 \\
        & w/o RBS & 0.2516 & 0.3061 & 0.2015 & 0.1537 & 0.2879 & 0.1552 \\
        & \textit{w/o Align}$^\ddagger$ & \textit{0.2722} & \textit{0.3089} & \textit{0.2411} & \textit{0.1454} & \textit{0.2967} & \textit{0.1716} \\
        \midrule
        \multirow{6}{*}{GDELT} & \cellcolor{grayinspire}\method{} & \cellcolor{grayinspire}0.1623 & \cellcolor{grayinspire}0.1072 & \cellcolor{grayinspire}0.2176 & \cellcolor{grayinspire}0.000134 & \cellcolor{grayinspire}0.1639 & \cellcolor{grayinspire}0.1272 \\
        & w/o CEW & 0.1928 & 0.1185 & 0.2674 & 0.000144 & 0.2112 & 0.1597 \\
        & w/o LRE & 0.1744 & 0.1175 & 0.2315 & 0.000153 & 0.1769 & 0.1376 \\
        & w/o CEW+LRE & 0.1721 & 0.1132 & 0.2313 & 0.000153 & 0.1750 & 0.1361 \\
        & w/o RBS & 0.1538 & 0.1061 & 0.2019 & 0.000157 & 0.1510 & 0.1187 \\
        & \textit{w/o Align}$^\ddagger$ & \textit{0.1653} & \textit{0.1074} & \textit{0.2236} & \textit{0.000133} & \textit{0.1699} & \textit{0.1311} \\
        \bottomrule
    \end{tabular*}
\end{table*}

\clearpage
\makeatletter\global\@colht\textheight\global\@colroom\textheight\makeatother

\begin{table*}[!t]
    \centering\small
    \renewcommand{\arraystretch}{0.95}
    \caption{Complete benchmark statistics under the chronological $50\%/20\%/30\%$ timestamp split. ``Train / Val / Test (eval)'' lists unique snapshot counts per split; following the released evaluation horizon, the last two tail test timestamps are excluded during evaluation (shown in parentheses). ``Avg.\ facts/snap'' reports total facts divided by total unique timestamps.}
    \label{tab:supp-dataset-stats}
    \begin{tabular*}{\textwidth}{@{\extracolsep{\fill}}lrrrrrr@{}}
        \toprule
        Dataset & Entities $|\mathcal{E}|$ & Relations $|\mathcal{R}|$ & Total facts $|\mathcal{F}|$ & Snapshots $|\mathcal{T}|$ & Train / Val / Test (eval) & Avg.\ facts/snap\\
        \midrule
        ICEWS14     & 7,128  & 230 & 90,730    & 365   & 182 / 73 / 110 (108)       & 248.6\\
        ICEWS18     & 23,033 & 256 & 468,558   & 304   & 152 / 60 / 92 (90)         & 1,541.3\\
        ICEWS05--15 & 10,488 & 251 & 461,329   & 4,017 & 2,008 / 803 / 1,206 (1,204) & 114.8\\
        GDELT       & 7,691  & 240 & 2,278,405 & 2,751 & 1,375 / 550 / 826 (824)    & 828.2\\
        \bottomrule
    \end{tabular*}
\end{table*}
\begin{table*}[!t]
    \centering\small
    \renewcommand{\arraystretch}{0.95}
    \caption{Entity emergence profile across four TKG benchmarks in the evaluated test intervals (following the released evaluation horizon). ``Share of test-active entities'' uses distinct test-active entities as denominator; ``Emerging-query share'' uses all directed test queries; ``Cohort queries'' track both debut and post-debut answer occurrences of test-emergent identities.}
    \label{tab:supp-emergence-profile}
    \begin{tabular*}{\textwidth}{@{\extracolsep{\fill}}lrrrrr@{}}
        \toprule
        \multicolumn{1}{c}{\raisebox{0.55ex}{Dataset}} & \shortstack[r]{Directed test\\queries} & \shortstack[r]{Test-emergent\\answer entities} & \shortstack[r]{Share of test-active\\entities} & \shortstack[r]{Emerging-query\\share} & \shortstack[r]{Cohort\\queries}\\
        \midrule
        ICEWS14     & 57,200    & 1,040 & 24.46\% & 4.54\%  & 2,200\\
        ICEWS18     & 257,558   & 2,692 & 20.52\% & 2.72\%  & 5,113\\
        ICEWS05--15 & 268,454   & 1,593 & 25.20\% & 1.49\%  & 12,249\\
        GDELT       & 1,554,472 & 785   & 13.33\% & 0.166\% & 2,080\\
        \bottomrule
    \end{tabular*}
\end{table*}
\begin{table*}[!t]
    \centering\small
    \renewcommand{\arraystretch}{0.95}
    \caption{Emerging-answer cohort denominators by natural support stratum, reported as directed queries / distinct answer identities ($q/e$).}
    \label{tab:supp-cohort-denominators}
\begin{tabular*}{\textwidth}{@{\extracolsep{\fill}}lrrrrr@{}}
\toprule
Dataset ($q/e$) & 0 & 1 & 2--5 & 6--10 & $>10$ \\
\midrule
ICEWS14 & 1,375/1,040 & 233/185 & 330/143 & 136/34 & 126/20 \\
ICEWS18 & 3,597/2,692 & 526/402 & 615/284 & 148/39 & 227/16 \\
ICEWS05--15 & 2,043/1,593 & 751/580 & 1,885/591 & 1,241/297 & 6,329/198 \\
GDELT & 1,294/785 & 183/137 & 400/158 & 120/32 & 83/9 \\
\bottomrule
\end{tabular*}

\end{table*}
\begin{table*}[!t]
    \centering\small
    \renewcommand{\arraystretch}{0.95}
    \setlength{\tabcolsep}{3pt}
    \caption{Independent Stage I/II training times, peak Stage II GPU memory, and scoring latency change at $p_H=p_C=0.2$. Times include validation and checkpoint I/O; scoring change compares median full timestamp latencies of \method{} vs.\ predictor.}
    \label{tab:efficiency-extended}
    \begin{tabular*}{\textwidth}{@{\extracolsep{\fill}}lrrrr@{}}
        \toprule
        Dataset & Stage I (s) & Stage II (s) & GPU (GiB) & Scoring change \\
        \midrule
        ICEWS14 & 2,931.1 & 683.5 & 2.50 & $+8.64\%$ to $+9.48\%$ \\
        ICEWS18 & 12,359.7 & 5,018.0 & 7.14 & $+4.84\%$ to $+6.07\%$ \\
        ICEWS05--15 & 20,772.7 & 5,716.5 & 2.17 & $+12.84\%$ to $+18.62\%$ \\
        GDELT & 18,882.8 & 11,619.8 & 8.24 & $-4.37\%$ to $+2.56\%$ \\
        \bottomrule
    \end{tabular*}
\end{table*}

\section{Dataset Statistics, Emergence Profile, and Costs}
\autoref{tab:supp-dataset-stats} summarizes the complete catalog sizes, total facts, unique timestamps, and $50\%/20\%/30\%$ chronological snapshot splits across all four benchmarks, complementing the test-interval emergence denominators in \autoref{tab:supp-emergence-profile} (corresponding to \autoref{fig:emergence-example}(c) of the main manuscript) and the support-stratum breakdown in \autoref{tab:supp-cohort-denominators}. Across the four benchmarks, training fact density averages $243.6$, $1{,}632.9$, $123.1$, and $736.7$ facts per snapshot on ICEWS14, ICEWS18, ICEWS05--15, and GDELT, respectively. Because an entity identity may accumulate support across timestamps and appear in multiple strata, overall $\operatorname{MRR}_{\mathrm{macro}}$ averages directly over distinct identities rather than over macro means of individual bins. Across all datasets, zero-support queries constitute a major share of the emerging-answer cohort, whereas ICEWS05--15 exhibits heavy post-debut traffic ($6{,}329$ queries over $198$ identities in the $>10$ bin).

\autoref{tab:efficiency-extended} summarizes the independent Stage I and Stage II computational costs at $p_H=p_C=0.2$. Stage II adapter training requires $23.3\%$--$61.5\%$ of Stage I wall-clock time across the four benchmarks, keeping sampled Stage II peak GPU memory between $2.17$ and $8.24$\,GiB while introducing only modest changes in scoring latency across three evaluation panels relative to the base predictor.

\section{Supporting Parameter and Diagnostic Protocols}
Supporting Stage II sensitivity evaluations use $(p_H=p_C=0.2)$ predictors on ICEWS14 and ICEWS18, testing queue caps $K\in\{16,32,64\}$ at $\gamma=0.5$ and correlation strengths $\gamma\in\{0,0.1,0.5,1,2\}$ at $K=32$ strictly on validation splits. In contrast, the main paper's intersecting masking sweeps train a dedicated Stage II adapter for each $(p_H,p_C)$ predictor.

Independent training costs include validation after each epoch and checkpoint I/O in separate processes on RTX 4090 GPUs. Stage I times cover initialization, validation, and interim tests through early stopping, whereas Stage II times cover direct adapter training, validation, and checkpoint writes. Inference scoring overhead compares median full timestamp latencies across five primary repeats (excluding sorting, with state feedback timed separately).
\clearpage

\section{Posterior Mean--Variance Coupling and Entity States}
\begin{figure*}[!b]
    \centering
    \includegraphics[width=0.90\textwidth]{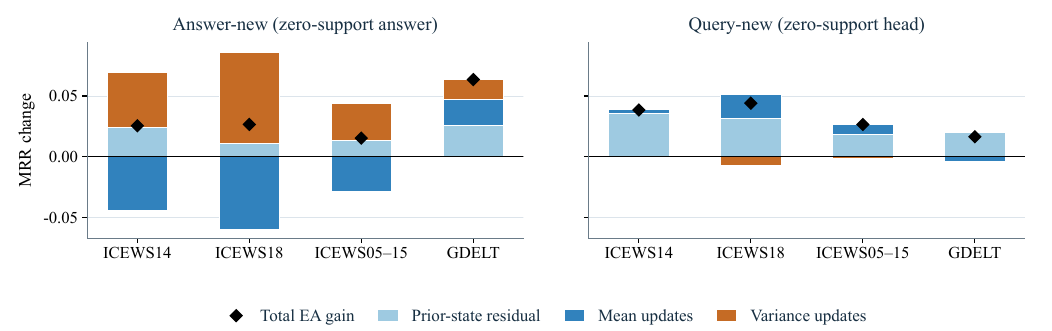}
    \caption{Complementary Shapley decomposition of posterior mean vs.\ variance updates across Answer-new (left) and Query-new (right) at fixed checkpoints $(\theta^\ast,\phi^\ast)$. Because mean and variance are coupled inside the Gaussian compatibility score of the main paper, updating established candidates' means while holding variance frozen shifts Answer-new negatively on ICEWS yet improves Query-new across all three ICEWS datasets (and Answer-new on GDELT), whereas strictly positive variance contraction protects Answer-new on all four datasets.}
    \label{fig:supp-mean-variance}

    \medskip
    \includegraphics[width=0.86\textwidth]{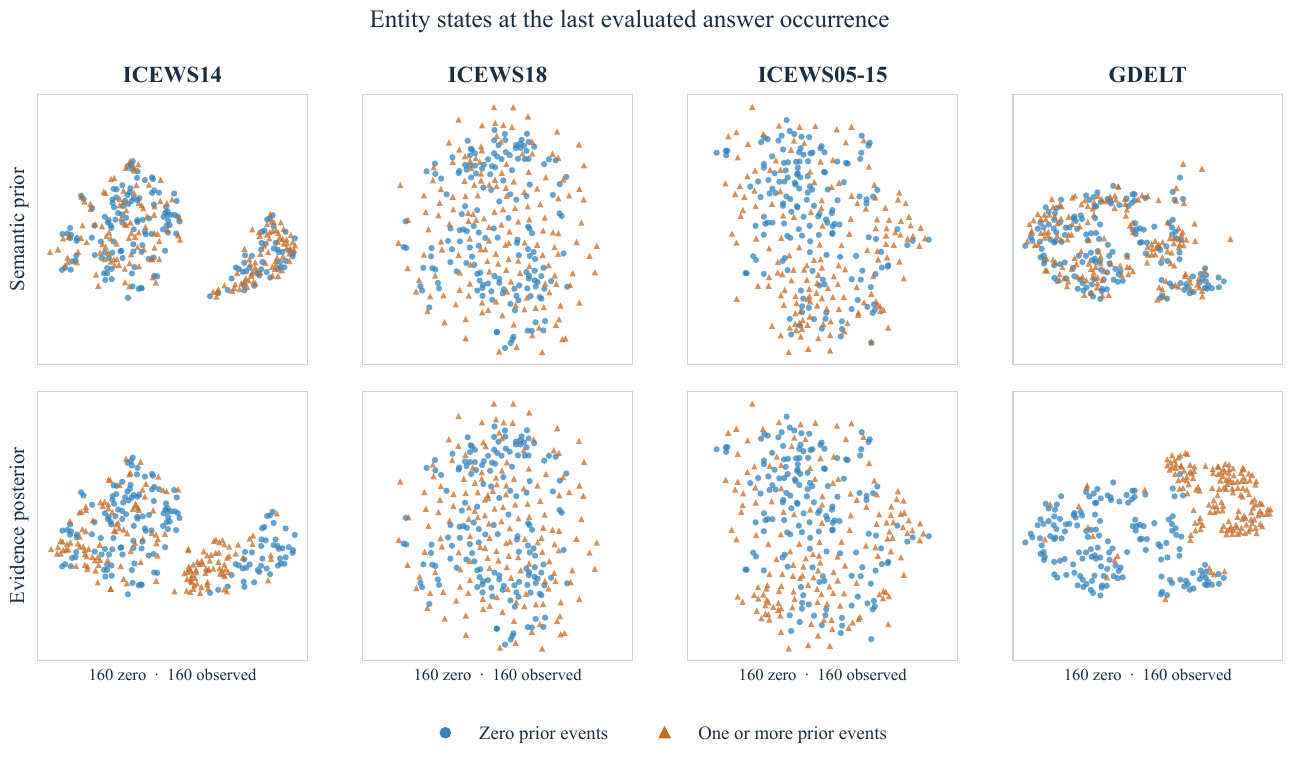}
    \caption{Joint t-SNE projections of 64-dimensional semantic prior means and evidence posterior means for sampled test-emergent answer identities (160 zero-event and 160 observed per dataset) at their last evaluated occurrence.}
    \label{fig:entity-state-map}
\end{figure*}

\autoref{fig:supp-mean-variance} presents a complementary Shapley decomposition of posterior mean and variance updates on Answer-new and Query-new queries at fixed trained checkpoints $(\theta^\ast,\phi^\ast)$, showing how strictly positive variance contraction protects Answer-new accuracy across all four benchmarks while mean updates drive Query-new gains on ICEWS.

\autoref{fig:entity-state-map} visualizes joint t-SNE projections of $64$-dimensional semantic prior means and evidence posterior means for sampled test-emergent answer identities at their last evaluated occurrence, illustrating how zero-event entities remain anchored to their semantic priors while observed entities migrate coherently with accumulated evidence.

\end{document}